\documentclass{article}

\usepackage{iclr2027_conference,times}

\usepackage{url}
\usepackage{amsmath}
\usepackage{amssymb}
\usepackage{mathtools}
\usepackage{graphicx}
\usepackage{booktabs}
\usepackage{multirow}
\usepackage{microtype}
\usepackage{array}
\usepackage{tabularx}
\usepackage{multicol}
\usepackage{colortbl}
\usepackage[table]{xcolor}
\usepackage{wrapfig}
\usepackage{makecell}
\usepackage{algorithm}
\usepackage{algpseudocode}
\usepackage{longtable}
\usepackage{xurl}

\usepackage{tikz}
\usetikzlibrary{arrows.meta, positioning, fit, backgrounds, calc}

\providecolor{iclrheader}{RGB}{240,244,248}
\providecolor{iclrsubtext}{RGB}{100,110,120}

\newcolumntype{Y}{>{\centering\arraybackslash}X}

\newcommand{\clsresult}[2]{%
  \mbox{#1\,/\,\textcolor{iclrsubtext}{#2}}%
}

\title{MedBenchAgent: Towards Systematic Automation of Medical VLM Benchmark Construction}

\author{
\parbox{0.94\textwidth}{\bfseries
\mbox{Yulin Fu$^{1,*}$},
\mbox{Junren Wang$^{2,3,*}$},
\mbox{Guangjing Yang$^{1}$},
\mbox{Zhangyuan Yu$^{1}$},
\mbox{Wanran Sun$^{1}$},
\mbox{Jiabao Zhou$^{1}$},
\mbox{Jin Yin$^{2,3}$},
\mbox{Qicheng Lao$^{1}$}
} \\[0.5em]
$^{1}$Beijing University of Posts and Telecommunications \\
$^{2}$West China Hospital \\
$^{3}$Sichuan Provincial Engineering Research Center of Intelligent Diagnosis and Treatment of \\
\hspace{1.2em}Breast Diseases \\[0.3em]
$^{*}$Equal contribution. \\
\texttt{\{yulin.fu, qicheng.lao\}@bupt.edu.cn}
}

\iclrfinalcopy

\newcounter{desideratum}
\renewcommand{\thedesideratum}{D\arabic{desideratum}}

\newcommand{\desideratum}[2]{%
  \refstepcounter{desideratum}%
  \par
  \noindent\textbf{\thedesideratum: #1.}%
  \label{#2}\enspace
}

\newcommand{\spresult}[2]{%
  \mbox{#1\,/\,\textcolor{iclrsubtext}{#2}}%
}

\usepackage{hyperref}
\definecolor{iclrlink}{RGB}{24, 112, 192}

\hypersetup{
    colorlinks=true,
    allcolors=iclrlink,
    linkbordercolor={1 1 1},
    pdfborder={0 0 0}
}

\begin{document}
\maketitle
\fancyhead{} 

\begin{abstract}
Large-scale construction of medical vision-language model (VLM) benchmarks is increasingly feasible with richly annotated imaging datasets and large language models (LLMs), yet existing automation largely focuses on generating evaluation items within predefined benchmark specifications. We study the broader problem of automatically deriving the specification itself: \emph{what} to evaluate, \emph{which} annotations support each task, and \emph{how} to translate this evidence into reliable evaluation items. We formulate benchmark construction as \emph{constrained compilation}, in which the benchmark specification is progressively derived from evaluation requirements, heterogeneous annotations, and medical knowledge. Based on this formulation, we introduce \textbf{MedBenchAgent}, a multi-agent framework with a \textbf{Benchmark Intermediate Representation (BIR)} that encodes task definitions, evidence mappings, evaluation protocols, and item specifications across construction stages. MedBenchAgent separates \emph{planning}, which derives and verifies the specification, from \emph{instantiation}, which constructs and audits items under the locked specification. MedBenchAgent achieves a Task-Space F1 of 90.9\%, outperforming direct task induction (79.2--80.0\%) and prior-guided induction (85.1\%); 994 of 1,000 sampled items from correctly identified tasks pass human audit. We further demonstrate portability to a specialized medical domain and evaluate twelve VLMs, revealing task- and setting-specific variation obscured by aggregate scores. These results establish constrained compilation as a scalable and auditable framework for medical VLM benchmark construction beyond question generation.
\end{abstract}

\section{Introduction}

Large-scale construction of medical vision-language model (VLM) benchmarks is increasingly feasible as richly annotated imaging datasets provide reusable supervision and large language models (LLMs) enable scalable generation of evaluation items \citep{zhang2023pmc,hu2024omnimedvqa,chen2024gmai}. Yet a benchmark is more than a collection of images and questions: it must specify \emph{what} capabilities are evaluated, \emph{which} source annotations validly support them, \emph{how} annotation evidence is transformed into reference answers, \emph{how} evaluation items are instantiated, and \emph{how} model responses are elicited and assessed. Collectively, we refer to these decisions as the \emph{benchmark specification}, which in turn determines the claims that the resulting evaluation can support. In medical imaging, establishing such a specification requires interpreting heterogeneous annotation schemas, units, terminology, and label semantics, making manual benchmark design time-consuming and expertise-intensive. Automating the specification itself, rather than only large-scale item generation, is therefore an important problem for scalable and reliable medical VLM evaluation.

Existing work approaches automated benchmark construction from two directions. Medical systems construct benchmarks from domain data, but operate within predefined clinical scopes, reporting structures, or question schemas \citep{ehrbench,liu2026automated}. General-purpose systems offer greater flexibility by using declarative or agentic workflows to translate evaluation requirements into benchmark construction \citep{ICLR2025_eb216114,benchmarkagent,butt2025benchagents}. However, they do not explicitly constrain medical benchmark specification by annotation-supported task validity, reference provenance, and medical knowledge. The remaining challenge, then, is to combine flexible benchmark specification with the semantic and evidential constraints imposed by evaluation requirements and heterogeneous medical data.

We address this challenge with \textbf{MedBenchAgent}, a multi-agent framework for systematic medical VLM benchmark construction. We cast benchmark construction as \emph{constrained compilation}, in which the benchmark specification is progressively established from an intended evaluation scope, heterogeneous source annotations, and medical knowledge, with each confirmed decision narrowing the space of subsequent construction. To make these constraints explicit and preserve them across stages, we introduce a \textbf{Benchmark Intermediate Representation (BIR)} that records the evaluation scope, task hierarchy, annotation evidence mappings, evaluation protocols, sample allocations, and item-construction specifications. Together, constrained compilation and the BIR provide an explicit mechanism for carrying semantic and evidential constraints from benchmark specification into downstream construction. MedBenchAgent operationalizes this formulation through a staged multi-agent workflow for medical benchmark construction: 1) planning resolves and verifies the specification against dataset annotations and medical references before locking the BIR; and 2) instantiation constructs and audits evaluation items under this fixed specification. In this way, MedBenchAgent extends automation from predefined medical scopes or question structures to benchmark specification itself, while preserving the annotation and medical constraints required for reliable construction.

This formulation enables empirical evaluation of both the benchmark-construction process and the benchmark produced by MedBenchAgent. In construction analysis, MedBenchAgent achieves a Task-Space F1 of 90.9\%, compared with 79.2--80.0\% for direct task induction and 85.1\% when the same task prior is provided to the direct model. For correctly identified tasks, 994 of 1,000 sampled items pass human audit. Together, these results suggest that open-ended decisions about \emph{what should be evaluated} are substantially more challenging than evidence mapping and item construction once benchmark semantics are constrained. We further instantiate the workflow in a specialized medical domain under a different annotation schema and evaluate twelve VLMs on the resulting benchmark. Model evaluation reveals substantial variation across tasks, imaging settings, and evaluation metrics that aggregate scores can obscure, reinforcing fine-grained benchmark specification as an important component of medical VLM evaluation.

Overall, this work makes three contributions: (1) \textbf{Conceptually}, we model the construction of medical VLM benchmarks and formulate it as constrained compilation; (2) \textbf{Methodologically}, we introduce a multi-agent MedBenchAgent framework and its Benchmark Intermediate Representation (BIR); and (3) \textbf{Empirically}, we derive insights into both the benchmark-construction process and medical VLM evaluation, while providing reusable benchmark artifacts.

\section{Related Work}
\label{sec:related-work}

\noindent{\textbf{Medical VLM benchmarks.}} 
Medical VLM benchmarks have progressed from manually authored question--answer datasets such as VQA-RAD and SLAKE \citep{lau2018dataset,liu2021slake}, to scalable reuse of image--caption and classification supervision in PMC-VQA and OmniMedVQA \citep{zhang2023pmc,hu2024omnimedvqa}, and more recently to task-structured evaluation over heterogeneous datasets in GMAI-MMBench and MMBU \citep{chen2024gmai,MMBU}.

\noindent{\textbf{Automated benchmark construction.}} 
General-purpose benchmark automation has increasingly adopted declarative or agentic construction, whereas medical systems have favored domain-specific, more deterministic workflows. AutoBencher constructs benchmarks from specified evaluation desiderata, BenchmarkAgent translates high-level evaluation requirements into construction procedures, and BenchAgents coordinates multiple agents across benchmark-creation stages \citep{ICLR2025_eb216114,benchmarkagent,butt2025benchagents}. In the medical domain, EHRBench uses a task-specific template pipeline to instantiate clinical QA from EHR data \citep{ehrbench}, while \citet{liu2026automated} packages schema- and report-grounded oncology VQA generation as a reusable agent skill.

\section{What Makes a Good Medical VLM Benchmark?}
\label{sec:good-medical-vlm-benchmark}

\subsection{Benchmark Formulation}
\label{sec:benchmark-formulation}

A medical VLM benchmark must specify both what is evaluated and how model performance is assessed. 
Building on medical benchmarking frameworks that distinguish the evaluation objective, challenge data, reference annotations, and assessment procedures \citep{karargyris2023federated,maier2020bias}, and following recent task-oriented medical VLM benchmarks, we additionally specify the task definitions and queries presented to the model for vision–language evaluation. We represent a benchmark by its intended evaluation scope $\mathcal{U}$, task space $\mathcal{T}$, evaluation-item set $\mathcal{I}$, and evaluation protocol $\mathcal{P}$.
A medical VLM benchmark is therefore defined as
\begin{equation}
\mathcal{B}=(\mathcal{U},\mathcal{T},\mathcal{I},\mathcal{P}),
\qquad
\mathcal{I}=\{\iota_i=(x_i,t_i,q_i,r_i)\}_{i=1}^{N},
\label{eq:benchmark}
\end{equation}
where $x_i$ is a medical imaging case, $t_i\in\mathcal{T}$ the evaluated task, $q_i$ the model-facing query, and $r_i$ the benchmark reference; $\mathcal{P}$ specifies each task's response convention and scoring rule.

Because each $r_i$ is a benchmark-level reference, not a raw annotation copied directly from a source dataset, its provenance and semantics must be recorded.
The CLAIM~\footnote{Checklist for Artificial Intelligence in Medical Imaging} guideline requires medical imaging studies to document the source, definition, and acquisition procedure of the reference standard \citep{tejani2024checklist}. We represent the source datasets as
\begin{equation}
\mathcal{D}=\{\mathcal{D}_m\}_{m=1}^{M},
\qquad
\mathcal{D}_m=(\mathcal{X}_m,\mathcal{A}_m,\Sigma_m),
\label{eq:source-datasets}
\end{equation}
where $\mathcal{X}_m$ is the set of imaging cases in dataset $m$, $\mathcal{A}_m$ is its annotation collection, and $\Sigma_m$ specifies the annotation schema, annotation unit, granularity, label space, and field semantics.
An annotation record $a\in\mathcal{A}_m$ represents evidence encoded under the dataset-specific specification $\Sigma_m$; it becomes a benchmark reference $r_i$ only after its semantics have been matched to a task $t_i$ and mapped into the common benchmark representation.

Benchmark construction must therefore determine not only whether the available annotations support a proposed task, but also how the relevant annotation evidence should be interpreted as its reference answer.
We therefore abstract benchmark construction as
\begin{equation}
F:(\mathcal{D},\mathcal{K},\mathcal{U})\mapsto\mathcal{B},
\label{eq:benchmark-construction}
\end{equation}
where $\mathcal{K}$ denotes the medical knowledge required to interpret clinical concepts and evaluation semantics. Once constructed, the benchmark can be applied to a model set $\mathcal{M}$ to produce empirical evidence $\mathcal{E}$: $(\mathcal{B},\mathcal{M})\longrightarrow\mathcal{E}$. 
This separates benchmark construction, which defines the evaluation object, from benchmark application, which produces evidence about model behavior under the specified scope and protocol. We use \emph{benchmark specification} to denote the pre-instantiation decisions that determine the evaluation scope, dataset-specific task definitions and granularity, supporting evidence and reference construction, evaluation protocol, sample allocation, and item-construction rules. The concrete item set $\mathcal{I}$ is produced by instantiating this specification over the source cases.

\subsection{Desiderata for Benchmark Construction}
\label{sec:desiderata}

The formulation above motivates three desiderata for medical VLM benchmark construction.

\desideratum{Evidence Groundedness}{des:evidence-groundedness}
Because benchmark construction maps source annotations to evaluation tasks and reference answers, annotation availability alone is insufficient to justify either output.
Evidence groundedness requires that each task be supported by the schema and semantics of its source annotations and that each item-level reference $r_i$ be entailed by the annotation evidence used to construct it. 
SLAKE uses physician-provided annotations, while EHRXQA uses templates and executable queries~\citep{liu2021slake,NEURIPS2023_0c007ebe}.

\desideratum{Medical Alignment}{des:clinical-alignment}
Annotation support establishes that a task is constructible, but does not determine how it should be expressed under the intended medical evaluation scope. Medical alignment requires task definitions, response spaces, and queries presented to the model to be consistent with established medical concepts and terminology, and not merely to verbalize labels to one dataset. GMAI-MMBench reflects this desideratum by organizing evaluation around clinically defined tasks~\citep{chen2024gmai}.

\desideratum{Evaluative Utility}{des:evaluative-utility}
The application $(\mathcal{B},\mathcal{M})\rightarrow\mathcal{E}$ further implies that evidence-grounded and medically aligned items do not necessarily yield informative evidence about model behavior. Evaluative utility requires the benchmark to support separate assessment of model capabilities and preserve interpretable differences across tasks and imaging settings. OmniMedVQA, MultiMedEval, and GMAI-MMBench exemplify this desideratum through evaluation across multiple datasets, tasks, modalities, and model families~\citep{hu2024omnimedvqa,pmlr-v250-royer24a,chen2024gmai}.



\subsection{Benchmark Construction as Constrained Compilation}
\label{sec:constrained-compilation}
We therefore formulate benchmark construction as constrained compilation: planning progressively establishes the benchmark specification, and each confirmed decision restricts the admissible choices of subsequent stages within the mapping $F$. The evaluation scope constrains the relevant task categories; annotation semantics determine which concrete prediction targets can be instantiated within them and which evidence can support their references; medical knowledge and evaluation requirements further constrain task expression and assessment. This process requires an explicit intermediate state that preserves confirmed decisions across stages. Section~\ref{sec:medbenchagent} instantiates this state as the BIR and locks it before item construction.

\begin{figure}[t]
    \centering
    \includegraphics[width=\linewidth]{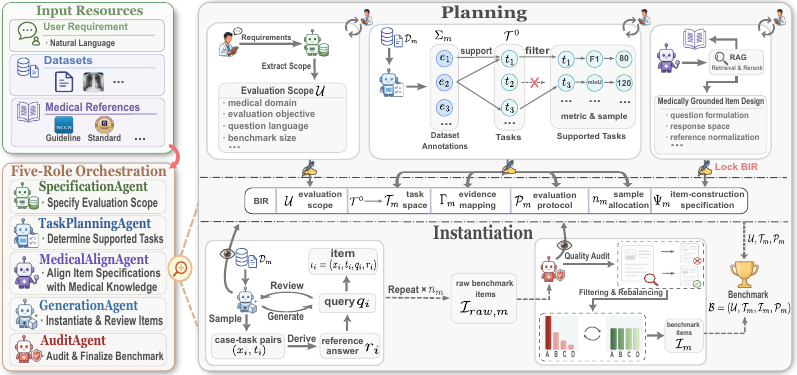}
    \setlength{\abovecaptionskip}{-4pt}
    \caption{
        \textbf{Overview of MedBenchAgent.} MedBenchAgent coordinates five agents through a shared Benchmark Intermediate Representation (BIR). Planning specifies tasks and their constraints within the BIR, whereas instantiation constructs evaluation items under the locked specifications.
        }
    \label{fig:method}
\end{figure}

\section{MedBenchAgent}
\label{sec:medbenchagent}

Figure~\ref{fig:method} summarizes the five-agent workflow and the separation between planning and instantiation, while Algorithm~\ref{alg:medbenchagent} formalizes the corresponding BIR updates and stage-wise construction procedure.

\subsection{Inputs and Benchmark Intermediate Representation}
\label{sec:inputs-and-bir}

MedBenchAgent takes three benchmark-specific resources as input:
a natural-language \textbf{user requirement}, \textbf{medical imaging datasets}, and \textbf{medical reference materials}. Each dataset-specific BIR contains a task-category prior $\mathcal{T}_0$, represented as an unexpanded task tree.
The user requirement $u$ defines the intended evaluation scope $\mathcal{U}$ and specifies additional user constraints on benchmark construction. The source collection $\mathcal{D}=\{\mathcal{D}_m\}_{m=1}^{M}$ consists of medical imaging cases and their dataset-specific annotations. The medical knowledge resource $\mathcal{K}$ consists of clinical guidelines, reporting standards, and other domain references relevant to the target setting. The initial task definitions in $\mathcal{T}_0$ are distilled from the task-design principles of existing medical VLM benchmarks and specify candidate tasks and their annotation-support conditions.

Section~\ref{sec:good-medical-vlm-benchmark} formulates benchmark construction as $F:(\mathcal{D},\mathcal{K},\mathcal{U})\mapsto\mathcal{B}$. MedBenchAgent implements this transformation through two coupled stages:
\begin{equation}
\mathcal{Z}
=
F_{\mathrm{plan}}
(\mathcal{D},\mathcal{K},\mathcal{U};\mathcal{T}_0),
\qquad
\mathcal{I}
=
F_{\mathrm{inst}}
(\mathcal{D},\mathcal{Z}),
\label{eq:medbenchagent-stages}
\end{equation}
where $\mathcal{Z}$ denotes the benchmark design produced during planning.

For each source dataset $\mathcal{D}_m$, planning maintains a dataset-specific \textbf{Benchmark Intermediate Representation (BIR)}. After the evaluation scope is confirmed, its task-space component is initialized as the unexpanded task tree $\mathcal{T}_0$. Candidate nodes are retained, removed, or expanded according to their support conditions, the evaluation scope, and available annotation evidence. Planning completes their evidence mappings, evaluation protocols, and sample allocations, while task design supplies the item-construction specifications. The resulting state is then confirmed and locked:
\begin{equation}
\mathrm{BIR}_m^{(0)}
=
\left(
\mathcal{U},
\mathcal{T}_0,
0,
0,
0,
0
\right),
\qquad
\mathrm{BIR}_m^{\mathrm{lock}}
=
\left(
\mathcal{U},
\mathcal{T}_m,
\Gamma_m,
\mathcal{P}_m,
n_m,
\Psi_m
\right).
\label{eq:bir}
\end{equation}
Here, $\mathcal{T}_m$ is the dataset-specific task tree obtained by retaining supported nodes and expanding decomposable tasks into distinct prediction targets. Ordinary tasks remain leaf nodes, whereas decomposable tasks have question types as their children. Each child must remain within its parent's task semantics and have annotation support for its own prediction target. 
The middle four components jointly describe this tree and its node specifications: $\Gamma_m(t)$ records supporting annotation evidence, $\mathcal{P}_m(t)$ specifies the evaluation protocol, and $n_m(t)$ denotes the sample allocation associated with task node $t$. These specifications are completed for the retained atomic tasks during planning; the zero entries in the initial BIR denote components not yet populated for the dataset. $\Psi_m(t)$ contains the corresponding medically grounded item-construction specification. We write $\operatorname{Leaves}(\mathcal{T}_m)$ for the retained atomic evaluation tasks, so a decomposable parent is not counted in addition to its children.

The planning output is therefore $\mathcal{Z}=\{\mathrm{BIR}_m^{\mathrm{lock}}\}_{m=1}^{M}$. Each $\mathrm{BIR}_m^{\mathrm{lock}}$ encodes the confirmed dataset-level benchmark specification, while instantiation produces the corresponding audited item set $\mathcal{I}_m$. Dataset-specific leaf tasks, item sets, and protocols are then consolidated across datasets, with $\mathcal{T}=\bigcup_{m=1}^{M}\operatorname{Leaves}(\mathcal{T}_m)$, to form the final benchmark $\mathcal{B}=(\mathcal{U},\mathcal{T},\mathcal{I},\mathcal{P})$.



\subsection{Planning: Compiling and Locking the BIR}
\label{sec:benchmark-planning}

Planning implements $F_{\mathrm{plan}}$ via SpecificationAgent, TaskPlanningAgent, and MedicalAlignAgent, which structure requirements, determine supported tasks, and design task items, respectively. The \emph{CKPT} markers in Algorithm~\ref{alg:medbenchagent} denote critical verification checkpoints at consequential planning transitions, validating the proposed BIR state before commitment through deterministic checks, human review, model-based supervision, or hybrid mechanisms.


\setlength{\intextsep}{4pt}

\begin{wrapfigure}{r}{0.60\textwidth}
\vspace{-0.4\baselineskip}

\begin{minipage}{\linewidth}
\scriptsize

\refstepcounter{algorithm}
\label{alg:medbenchagent}

\hrule
\vspace{2pt}

\noindent
\textbf{Algorithm \thealgorithm}\quad
Medical VLM benchmark construction

\vspace{2pt}
\hrule
\vspace{2pt}

\begin{algorithmic}[1]

\Require User requirement $u$, source datasets $\mathcal{D}$,
medical references $\mathcal{K}$

\Ensure Benchmark
$\mathcal{B}=(\mathcal{U},\mathcal{T},\mathcal{I},\mathcal{P})$

\State
$\mathcal{U}\xleftarrow{\mathrm{CKPT}}\Call{Specify}{u}$;
\quad
$\{\mathrm{BIR}_m^{(0)}
=
(\mathcal{U},\mathcal{T}_0,0,0,0,0)\}_{m=1}^{M}$;
\quad
$\mathcal{I},\mathcal{I}_{\mathrm{raw},m}\gets\emptyset,\ \forall m$

\ForAll{$\mathcal{D}_m\in\mathcal{D}$}

    \State
    $(\mathcal{T}_m,\Gamma_m,\mathcal{P}_m,n_m)
    \xleftarrow{\mathrm{CKPT}}
    \Call{AssessTaskSupport}
    {\mathcal{D}_m,\mathrm{BIR}_m^{(0)}}$

    \State
    $\mathrm{BIR}_m
    \gets
    (\mathcal{U},
     \mathcal{T}_m,
     \Gamma_m,
     \mathcal{P}_m,
     n_m,
     0)$

    \State
    $\Psi_m
    \xleftarrow{\mathrm{CKPT}}
    \Call{DesignTaskItems}
    {\mathcal{K},
     \mathcal{U},
     \mathcal{T}_m,
     \Gamma_m,
     \mathcal{P}_m}$

    \State
    $\mathrm{BIR}_m^{\mathrm{lock}}
    \gets
    (\mathcal{U},
     \mathcal{T}_m,
     \Gamma_m,
     \mathcal{P}_m,
     n_m,
     \Psi_m)$

    \ForAll{eligible case--task pairs $(x_i,t)$ with
    $t\in\operatorname{Leaves}(\mathcal{T}_m)$}

        \State
        $e_i\gets
        \Call{ExtractEvidence}{x_i,\Gamma_m(t)}$

        \State
        $r_i\gets
        \Call{DeriveReference}{e_i,\Psi_m(t)}$

        \State
        $q_i\gets
        \Call{GenerateQuery}{x_i,t,e_i,\Psi_m(t)}$

        \State
        $\iota_i\gets(x_i,t,q_i,r_i)$

        \State \textbf{if}
        $\Call{ReviewItem}
        {\iota_i,e_i,\mathrm{BIR}_m^{\mathrm{lock}}}$
        \textbf{ then }
        $\mathcal{I}_{\mathrm{raw},m}
        \gets
        \mathcal{I}_{\mathrm{raw},m}\cup\{\iota_i\}$

    \EndFor

    \State
    $\mathcal{I}_m
    \gets
    \Call{Audit}
    {\mathcal{I}_{\mathrm{raw},m},
     \mathrm{BIR}_m^{\mathrm{lock}},
     \mathcal{D}_m}$

    \State
    $\mathcal{I}
    \gets
    \mathcal{I}\cup\mathcal{I}_m$

\EndFor

\State
$\mathcal{T}\gets
\displaystyle\bigcup_{m=1}^{M}\operatorname{Leaves}(\mathcal{T}_m)$;
\quad
$\mathcal{P}\gets
\Call{MergeProtocols}
{\{\mathcal{P}_m\}_{m=1}^{M}}$

\State
\Return
$\mathcal{B}
=
(\mathcal{U},
 \mathcal{T},
 \mathcal{I},
 \mathcal{P})$

\end{algorithmic}

\vspace{2pt}
\hrule

\end{minipage}

\vspace{-0.4\baselineskip}
\end{wrapfigure}
\textbf{SpecificationAgent} first converts the open-ended natural-language requirement $u$ into a structured evaluation scope $\mathcal{U}$. The specification resolves ambiguous dimensions of the requested benchmark before constraining subsequent planning.

\textbf{TaskPlanningAgent} then extends the task-category prior into the dataset-specific task tree $\mathcal{T}_m$ using the evaluation scope and annotation evidence in $\mathcal{D}_m$. A category is retained only when its support conditions are satisfied, and $\Gamma_m$ records the eligible annotation fields to construct references for its tasks. When annotations expose multiple distinct prediction targets within a supported task, the agent expands it into child tasks, each representing an annotation-supported answer axis within parent semantics. Semantic duplicates are removed across retained atomic tasks, establishing the task support required by Desideratum~\ref{des:evidence-groundedness}. TaskPlanningAgent selects a compatible primary evaluation metric for each retained atomic task. The resulting task plan passes through CKPT before commitment to the BIR, after which deterministic allocation assigns $n_m$. Preserving prediction targets provides the basis for separate task-level analysis under Desideratum~\ref{des:evaluative-utility}.

\textbf{MedicalAlignAgent} subsequently converts each retained task into a medically meaningful and executable item-construction specification $\Psi_m$. It adopts retrieval-augmented generation (RAG)~\citep{rag} over the relevant medical references in $\mathcal{K}$ to ground the design in medical knowledge. Grounded on these references and constrained by the retained task semantics and authorized annotation evidence, it specifies the task's question semantics, response space, and reusable question formulations, together with the normalization to benchmark-level reference answers. This step connects an evidence-supported task to its medical evaluation semantics, thereby supporting Desideratum~\ref{des:clinical-alignment}. After the complete set of task designs passes through CKPT, the resulting
$\mathrm{BIR}_m^{\mathrm{lock}}$ is fixed for downstream instantiation. Downstream stages may read the confirmed task definitions, supporting evidence, protocols, sample allocations, and construction specifications, but cannot revise them.

\begin{figure}[t]
    \centering
    \includegraphics[width=\linewidth]{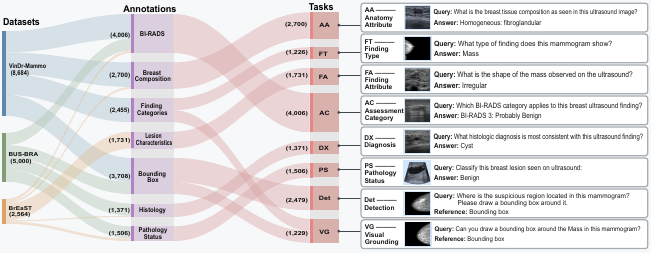}
    \setlength{\abovecaptionskip}{-8pt}
    \caption{\textbf{Benchmark construction and representative items.} Three source datasets are mapped through heterogeneous annotations to the retained evaluation tasks, with one representative item shown for each task. Numbers in parentheses denote item counts.}
    \label{fig:benchmark}
\end{figure}
\subsection{Instantiation: Constructing Items under the Locked BIR}
\label{sec:benchmark-instantiation}
Instantiation realizes $F_{\mathrm{inst}}$ in
Eq.~\ref{eq:medbenchagent-stages} by applying
$\mathrm{BIR}_m^{\mathrm{lock}}$ to the source annotations without
reopening the confirmed benchmark design. As summarized in Algorithm~\ref{alg:medbenchagent}, GenerationAgent operates at the case--task level to instantiate evaluation items, whereas AuditAgent operates over the resulting item set to verify benchmark-level consistency.

For each eligible case--task pair with $t_i\in\operatorname{Leaves}(\mathcal{T}_m)$, \textbf{GenerationAgent} constructs an evaluation item $\iota_i=(x_i,t_i,q_i,r_i)$ under the corresponding task specification $\Psi_m(t_i)$. The model generates only the query $q_i$, varying its wording within the locked task semantics and authorized annotation evidence. Deterministic code derives the reference answer $r_i$ from the supporting evidence specified by $\Gamma_m(t_i)$ and, where applicable, organizes the response options according to the confirmed construction specification. A subsequent model review call screens the generated item for task drift, ambiguity, unsupported additions, and answer leakage without modifying the fixed reference. This separation allows linguistic variation in the query while preserving annotation-grounded reference answers and the item-level entailment requirement of Desideratum~\ref{des:evidence-groundedness}.

\textbf{AuditAgent} subsequently checks the generated item set for consistency with the locked task-node specifications, source annotations, and benchmark-level requirements. Deterministic checks verify annotation support, response and reference consistency, and exact duplication, while accepted multiple-choice items are balanced with respect to option positions. AuditAgent additionally performs task-stratified consistency screening to identify issues such as task drift, answer leakage, ambiguity, and semantic duplication. The items passing the deterministic audit constitute the finalized evaluation-item set $\mathcal{I}$ and complete the resulting benchmark $\mathcal{B}=(\mathcal{U},\mathcal{T},\mathcal{I},\mathcal{P})$.

\section{Experiments: Benchmark Construction and Evaluation}
\label{sec:experiments}
\subsection{Experimental Setup}
\label{sec:experimental-setup}
\begin{table*}[t]
\centering

\begin{minipage}[t]{0.58\textwidth}
\centering
\scriptsize

\setlength{\tabcolsep}{1.5pt}

\renewcommand{\arraystretch}{1.05}

\setlength{\aboverulesep}{0pt}
\setlength{\belowrulesep}{0pt}

\caption{\textbf{Task-space construction and evidence-mapping results.}
Task-Space Precision (P), Recall (R), and F1 measure agreement with researcher review;
EMA measures agreement of supporting annotation fields (\%).}
\label{tab:task_design_review}

\vspace{3pt}

%
%
%

\begin{tabularx}{\linewidth}{
>{\raggedright\arraybackslash
  \hsize=2.3\hsize\linewidth=\hsize}X
*{3}{
>{\centering\arraybackslash
  \hsize=0.5\hsize\linewidth=\hsize}X
}
*{3}{
>{\centering\arraybackslash
  \hsize=0.5\hsize\linewidth=\hsize}X
}
!{\color{iclrsubtext}\vrule width 0.45pt}
>{\centering\arraybackslash
  \hsize=1\hsize\linewidth=\hsize}X
}

\toprule


\rowcolor{iclrheader}
&
\multicolumn{6}{c}{
    \cellcolor{iclrheader}\textbf{Task-Space Review}
}
&
\cellcolor{iclrheader}\textbf{Evidence}
\\

\noalign{%
    {\color{iclrheader}\hrule height \cmidrulewidth}%
    \vskip-\cmidrulewidth
}

\cmidrule(lr){2-7}
\cmidrule(lr){8-8}


\rowcolor{iclrheader}
\multicolumn{1}{c}{
    \multirow{-2}{*}{\textbf{Model}}
}
&
\textbf{TP}
&
\textbf{FP}
&
\textbf{FN}
&
\textbf{P$\uparrow$}
&
\textbf{R$\uparrow$}
&
\textbf{F1$\uparrow$}
&
\textbf{EMA$\uparrow$}
\\

\midrule

DeepSeek-V4-Flash
& 19
& 8
& 2
& 70.4
& 90.5
& 79.2
& 100.0
\\

Qwen3.5-9B
& 19
& 8
& 2
& 70.4
& 90.5
& 79.2
& 100.0
\\

MedGemma-27B
& 20
& 9
& 1
& 69.0
& 95.2
& 80.0
& 100.0
\\

DeepSeek-V4-Flash + $\mathcal{T}_0$
& 20
& 6
& 1
& 76.9
& 95.2
& 85.1
& 100.0

\\

\midrule

\textbf{MedBenchAgent}
& \textbf{20}
& \textbf{3}
& \textbf{1}
& \textbf{87.0}
& \textbf{95.2}
& \textbf{90.9}
& \textbf{100.0}
\\

\bottomrule

\end{tabularx}
\end{minipage}
\hfill
\begin{minipage}[t]{0.40\textwidth}
\centering
\scriptsize


\setlength{\tabcolsep}{2.0pt}

\renewcommand{\arraystretch}{1.20}

\setlength{\aboverulesep}{0pt}
\setlength{\belowrulesep}{0pt}

\caption{\textbf{Human consistency audit of finalized items.}
Pass rates are computed over 1,000 items sampled uniformly from the finalized benchmark.}
\label{tab:item_audit}

\vspace{5pt}

%
%

\begin{tabularx}{\linewidth}{
>{\raggedright\arraybackslash
  \hsize=2.0\hsize\linewidth=\hsize}X
>{\centering\arraybackslash
  \hsize=0.65\hsize\linewidth=\hsize}X
>{\centering\arraybackslash
  \hsize=0.55\hsize\linewidth=\hsize}X
>{\centering\arraybackslash
  \hsize=0.8\hsize\linewidth=\hsize}X
}

\toprule

\rowcolor{iclrheader}
\textbf{Criterion}
& \textbf{Pass}
& \textbf{Fail}
& \textbf{Pass Rate}
\\

\midrule

Reference groundedness
& 1,000
& 0
& 100.0\%
\\

Task--query alignment
& 996
& 4
& 99.6\%
\\

Unambiguity
& 1,000
& 0
& 100.0\%
\\

Evaluation integrity
& 994
& 6
& 99.4\%
\\

\midrule

\textbf{All criteria}
& \textbf{994}
& \textbf{6}
& \textbf{99.4\%}
\\

\bottomrule

\end{tabularx}
\end{minipage}

\end{table*}
\noindent{\textbf{Construction.}}
Figure~\ref{fig:benchmark} summarizes the resulting benchmark construction. We instantiate MedBenchAgent using three medical imaging datasets: VinDr-Mammo \citep{nguyen2023vindr}, BrEaST \citep{pawlowska2024curated}, and BUS-BRA \citep{gomez2024bus}. The medical knowledge resource is grounded in the relevant sections of the ACR BI-RADS v2025 Manual \citep{destounis2025mammography,leung2025ultrasound}. For the reported runs, we instantiate the planning checkpoints with researcher review. 

\noindent{\textbf{Evaluation.}} We evaluate twelve VLMs spanning general-purpose and medical-domain model families, multiple parameter scales, and different inference regimes. The general-purpose models include InternVL3-8B~\citep{internvl3}, Qwen2.5-VL (7B, 32B)~\citep{qwen25vl}, and Qwen3-VL (8B, 32B), including an 8B thinking variant~\citep{qwen3vl}. The medical models include MedGemma-4B~\citep{medgemma}, Lingshu (7B, 32B)~\citep{lingshu}, Hulu-Med-7B~\citep{hulumed}, QoQ-Med-7B~\citep{qoqmed}, and BreastGPT-8B~\citep{breastgpt}. Multiple-choice tasks are evaluated with Accuracy (Acc.) and Macro-F1 (F1), while spatial tasks use mean intersection over union (mIoU) and $\mathrm{Acc@IoU}_{0.5}$. When a task contains multiple fine-grained tasks, metrics are computed separately and macro-averaged with equal weight.

\noindent{\textbf{Implementation.}}
We host DeepSeek-V4-Flash \citep{DBLP:journals/corr/abs-2606-19348} as the language-model backend with vLLM \citep{kwon2023efficient}, and Qwen3-Embedding-0.6B and Qwen3-Reranker-0.6B \citep{zhang2025qwen3embeddingadvancingtext} for retrieval and reranking. Model evaluation uses deterministic decoding and FlashAttention-2~\citep{flash-attn}, and all experiments are run on NVIDIA B200 GPUs. 

\subsection{Benchmark Construction Analysis}
\label{sec:construction-analysis}

\noindent\textbf{Task-space construction.}
MedBenchAgent uses BIR-mediated task planning to constrain task-space construction with $\mathcal{U}$ and $\Sigma_m$, rather than inducing the task space in a single open-ended step. We evaluate this strategy against direct baselines using the same inputs in a single model call, and additionally provide DeepSeek-V4-Flash with the initial task definitions $\mathcal{T}_0$. The direct baselines use MedGemma-27B~\citep{medgemma}, Qwen3.5-9B~\citep{qwen35}, and DeepSeek-V4-Flash. We evaluate against a reference task space $\mathcal{T}^\star$, defined before MedBenchAgent and finalized prior to these experiments by dataset-familiar researchers with informal input from a clinical collaborator. Proposals are matched to $\mathcal{T}^\star$ by prediction target and annotation semantics rather than surface wording. We report Task-Space Precision, Recall, and F1 from TP, FP, and FN counts; for matched tasks, Evidence-Mapping Agreement (EMA) measures agreement with researcher-confirmed supporting fields. As shown in Table~\ref{tab:task_design_review}, BIR-mediated construction achieves 90.9\% Task-Space F1, compared with 79.2--80.0\% for direct induction. Providing $\mathcal{T}_0$ improves direct DeepSeek-V4-Flash induction to 85.1\% F1, while MedBenchAgent further increases precision from 76.9\% to 87.0\% at the same 95.2\% recall, reducing false positives from six to three. All methods achieve 100\% EMA on matched tasks. These results show that the BIR-mediated design primarily improves the selection of valid prediction targets, while supporting-field identification is reliable once the task is correctly identified.

\noindent\textbf{Human audit of finalized items.}
MedBenchAgent promotes consistent item construction by locking item-construction requirements in the confirmed specification before instantiation. To assess whether these constraints are preserved in the finalized benchmark, we sample 1,000 items without replacement and audit each for reference groundedness, task--query alignment, unambiguity, and evaluation integrity against its annotation evidence and specification. An item passes only when all four criteria are satisfied. Table~\ref{tab:item_audit} shows that 994 items pass, yielding a 99.4\% consistency rate. Six failures include four VinDr-Mammo finding-type questions that ask for BI-RADS assessment categories and two BrEaST diagnosis questions with unsupported additions. These cases localize remaining errors to template semantics and evidence control during generation, distinguishing item-expression errors from the task-specification errors identified in the task-space analysis. The high consistency rate supports locked specifications for large-scale item construction while identifying targets for semantic verification. Detailed setups and additional analyses are provided in Appendix~\ref{app:construction}.

\begin{table*}[t]
\centering
\scriptsize

\setlength{\tabcolsep}{3.6pt}
\renewcommand{\arraystretch}{1.30}

\setlength{\aboverulesep}{0pt}
\setlength{\belowrulesep}{0pt}

\caption{\textbf{Evaluation performance on tasks with choices.} Results are reported as \textbf{Accuracy} /
\textcolor{iclrsubtext}{\textbf{Macro-F1}} (\%).
$\dagger$ denotes thinking variants. Task abbreviations (AA, FT, etc.) are defined in Figure~\ref{fig:benchmark}.}
\label{tab:main_results}

\vspace{3pt}

\begin{tabularx}{\linewidth}{
>{\raggedright\arraybackslash}p{0.18\linewidth}
*{7}{Y}
}

\toprule


\rowcolor{iclrheader}
&
\textbf{Anatomy}
&
\multicolumn{2}{c}{
    \cellcolor{iclrheader}\textbf{Finding}
}
&
\multicolumn{3}{c}{
    \cellcolor{iclrheader}\textbf{Assessment \& Diagnosis}
}
&
\\

\noalign{%
    {\color{iclrheader}\hrule height \cmidrulewidth}%
    \vskip-\cmidrulewidth
}

\cmidrule(lr){2-2}
\cmidrule(lr){3-4}
\cmidrule(lr){5-7}


\rowcolor{iclrheader}
\multirow{-2}{*}{\textbf{Model}}
&
\textbf{AA}
&
\textbf{FT}
&
\textbf{FA}
&
\textbf{AC}
&
\textbf{DX}
&
\textbf{PS}
&
\multirow{-2}{*}{\textbf{Overall}}
\\

\midrule


\rowcolor{gray!10}
\multicolumn{8}{l}{
    \textbf{\textit{General-purpose VLMs}}
}
\\

InternVL3-8B
& \clsresult{36.59}{17.30}
& \clsresult{19.09}{8.65}
& \clsresult{42.28}{30.86}
& \clsresult{16.12}{12.82}
& \clsresult{35.35}{22.86}
& \clsresult{48.95}{40.25}
& \clsresult{33.06}{22.12}
\\

Qwen2.5-VL-7B
& \clsresult{48.86}{23.08}
& \clsresult{29.77}{10.17}
& \clsresult{41.45}{29.28}
& \clsresult{21.35}{11.93}
& \clsresult{33.76}{22.67}
& \clsresult{51.03}{41.86}
& \clsresult{37.70}{23.16}
\\

Qwen3-VL-8B-I
& \clsresult{34.89}{18.56}
& \clsresult{16.31}{9.03}
& \clsresult{29.84}{22.57}
& \clsresult{39.11}{19.74}
& \clsresult{39.98}{29.65}
& \clsresult{60.55}{47.85}
& \clsresult{36.78}{24.57}
\\

Qwen3-VL-8B-T$^\dagger$
& \clsresult{25.52}{14.18}
& \clsresult{15.09}{7.29}
& \clsresult{56.93}{35.13}
& \clsresult{38.58}{20.03}
& \clsresult{35.13}{24.15}
& \clsresult{53.05}{42.52}
& \clsresult{37.38}{23.88}
\\

Qwen2.5-VL-32B
& \clsresult{44.68}{21.59}
& \clsresult{54.00}{12.67}
& \clsresult{46.20}{29.55}
& \clsresult{26.83}{16.79}
& \clsresult{43.26}{26.69}
& \clsresult{52.92}{42.41}
& \clsresult{44.65}{24.95}
\\

Qwen3-VL-32B
& \clsresult{37.31}{20.47}
& \clsresult{\textbf{54.65}}{13.25}
& \clsresult{37.08}{28.42}
& \clsresult{34.22}{20.59}
& \clsresult{39.37}{29.83}
& \clsresult{59.37}{54.28}
& \clsresult{43.67}{27.81}
\\

\midrule


\rowcolor{gray!10}
\multicolumn{8}{l}{
    \textbf{\textit{Medical VLMs}}
}
\\

MedGemma-4B
& \clsresult{22.39}{15.69}
& \clsresult{23.16}{10.84}
& \clsresult{47.55}{33.94}
& \clsresult{17.98}{15.38}
& \clsresult{30.20}{23.75}
& \clsresult{55.78}{44.32}
& \clsresult{32.84}{23.99}
\\

Lingshu-7B
& \clsresult{44.27}{26.13}
& \clsresult{32.22}{12.28}
& \clsresult{56.84}{38.66}
& \clsresult{25.28}{20.85}
& \clsresult{52.18}{29.29}
& \clsresult{71.21}{52.53}
& \clsresult{47.00}{29.96}
\\

Hulu-Med-7B
& \clsresult{31.62}{16.63}
& \clsresult{17.05}{9.19}
& \clsresult{51.14}{35.49}
& \clsresult{22.75}{18.81}
& \clsresult{49.64}{34.70}
& \clsresult{64.14}{51.59}
& \clsresult{39.39}{27.73}
\\

QoQ-Med-7B$^\dagger$
& \clsresult{48.56}{22.72}
& \clsresult{53.83}{12.03}
& \clsresult{66.09}{37.85}
& \clsresult{20.43}{10.24}
& \clsresult{41.45}{24.07}
& \clsresult{42.14}{30.44}
& \clsresult{45.42}{22.89}
\\

BreastGPT-8B
& \clsresult{36.03}{\textbf{26.41}}
& \clsresult{35.73}{\textbf{16.96}}
& \clsresult{\textbf{75.22}}{\textbf{52.50}}
& \clsresult{\textbf{40.72}}{\textbf{31.54}}
& \clsresult{\textbf{65.09}}{\textbf{47.06}}
& \clsresult{\textbf{78.58}}{\textbf{71.30}}
& \clsresult{\textbf{55.23}}{\textbf{40.96}}
\\

Lingshu-32B
& \clsresult{\textbf{50.89}}{24.70}
& \clsresult{42.66}{14.17}
& \clsresult{57.54}{43.01}
& \clsresult{34.19}{21.65}
& \clsresult{49.04}{24.54}
& \clsresult{70.08}{55.25}
& \clsresult{50.73}{30.55}
\\


\midrule

\rowcolor{gray!10}
\multicolumn{8}{l}{
    \textbf{\textit{Family-level Summary}}
}
\\

\rowcolor{gray!5}
\textbf{General-purpose Avg.}
& \clsresult{37.98}{19.19}
& \clsresult{31.49}{10.18}
& \clsresult{42.30}{29.30}
& \clsresult{29.37}{16.98}
& \clsresult{37.81}{25.97}
& \clsresult{54.31}{44.86}
& \clsresult{38.87}{24.41}
\\

\rowcolor{gray!5}
\textbf{Medical Avg.}
& \clsresult{38.96}{22.04}
& \clsresult{34.11}{12.58}
& \clsresult{59.06}{40.24}
& \clsresult{26.89}{19.74}
& \clsresult{47.93}{30.57}
& \clsresult{63.65}{50.90}
& \clsresult{45.10}{29.35}
\\

\bottomrule

\end{tabularx}
\end{table*}

\subsection{Model Evaluation}
\label{sec:model_evaluation}

\setlength{\intextsep}{-9pt}

\begin{wraptable}{r}{0.48\columnwidth}
\centering
\scriptsize

\setlength{\tabcolsep}{0.6pt}
\renewcommand{\arraystretch}{1.05}
\setlength{\aboverulesep}{0pt}
\setlength{\belowrulesep}{0pt}

\caption{\textbf{Spatial localization performance.}
Results are reported as
\textcolor{iclrsubtext}{mIoU} /
$\mathrm{Acc@IoU}_{0.5}$ (\%).}
\label{tab:spatial_results}

\vspace{3pt}

\begin{tabularx}{\linewidth}{
>{\raggedright\arraybackslash}p{0.32\linewidth}
*{3}{Y}
}

\toprule


\rowcolor{iclrheader}
&
\multicolumn{2}{c}{
    \cellcolor{iclrheader}\textbf{VinDr-Mammo}
}
&
\cellcolor{iclrheader}\textbf{BUS-BRA}
\\

\noalign{%
    {\color{iclrheader}\hrule height \cmidrulewidth}%
    \vskip-\cmidrulewidth
}

\cmidrule(lr){2-3}
\cmidrule(lr){4-4}


\rowcolor{iclrheader}
\multirow{-2}{*}{\textbf{Model}}
&
\textbf{Det}
&
\textbf{VG}
&
\textbf{Det}
\\

\midrule


\rowcolor{gray!10}
\multicolumn{4}{l}{
    \textbf{\textit{General-purpose VLMs}}
}
\\

InternVL3-8B
& \spresult{3.30}{0.24}
& \spresult{3.15}{0.33}
& \spresult{26.00}{11.20}
\\

Qwen2.5-VL-7B
& \spresult{7.60}{3.91}
& \spresult{8.77}{6.51}
& \spresult{28.94}{22.24}
\\

Qwen3-VL-8B-I
& \spresult{10.72}{7.00}
& \spresult{13.68}{10.98}
& \spresult{51.11}{56.40}
\\

Qwen3-VL-8B-T$^\dagger$
& \spresult{10.94}{7.32}
& \spresult{10.42}{7.00}
& \spresult{37.50}{32.88}
\\

Qwen2.5-VL-32B
& \spresult{9.70}{6.10}
& \spresult{10.83}{7.32}
& \spresult{40.63}{37.52}
\\

Qwen3-VL-32B
& \spresult{\textbf{15.40}}{\textbf{12.86}}
& \spresult{\textbf{18.01}}{\textbf{16.84}}
& \spresult{55.36}{62.08}
\\

\midrule


\rowcolor{gray!10}
\multicolumn{4}{l}{
    \textbf{\textit{Medical VLMs}}
}
\\

MedGemma-4B
& \spresult{1.54}{0.00}
& \spresult{1.57}{0.00}
& \spresult{15.73}{3.68}
\\

Lingshu-7B
& \spresult{4.41}{0.81}
& \spresult{3.65}{0.65}
& \spresult{4.71}{0.88}
\\

Hulu-Med-7B
& \spresult{0.01}{0.00}
& \spresult{0.00}{0.00}
& \spresult{6.97}{2.96}
\\

QoQ-Med-7B$^\dagger$
& \spresult{6.97}{1.06}
& \spresult{6.97}{1.30}
& \spresult{45.63}{43.52}
\\

BreastGPT-8B
& \spresult{5.34}{0.90}
& \spresult{5.44}{1.14}
& \spresult{\textbf{72.62}}{\textbf{82.80}}
\\

Lingshu-32B
& \spresult{4.70}{1.46}
& \spresult{5.62}{1.95}
& \spresult{12.46}{4.40}
\\

\bottomrule

\end{tabularx}
\end{wraptable}


\textbf{Task-oriented specification preserves non-redundant evaluation evidence.}
\label{sec:mainresults}
MedBenchAgent retains annotation-supported prediction targets as separate tasks and records their evaluation protocols in the BIR. We test whether these choices affect model assessment beyond increasing benchmark granularity. As shown in Table~\ref{tab:main_results}, medical VLMs outperform general-purpose VLMs by 6.23 points in Overall Acc. but trail by 2.48 points on AC, indicating that the aggregate advantage is not uniform across tasks. To quantify such changes, we compare all 66 model pairs and find that 12--24 pairs (18.2--36.4\%) reverse ordering between Overall Acc. and individual-task Acc.; within tasks, 4--15 pairs (6.1--22.7\%) reverse between Acc. and F1.

MedBenchAgent also preserves localization annotations as spatial prediction targets rather than converting them into choice-based questions. For example, Lingshu-7B has higher Overall Acc. than Qwen2.5-VL-32B (47.00\% vs.\ 44.65\%), while Qwen2.5-VL-32B performs better on all six spatial measures in Table~\ref{tab:spatial_results}. Across these measures, 28--34 of 66 model pairs (42.4--51.5\%) reverse ordering relative to Overall Acc. Thus, rankings vary substantially across evaluation forms. These results support Desideratum~\ref{des:evaluative-utility}: preserving annotation-supported tasks across capability dimensions and assigning task-appropriate metrics materially changes the evidence provided by the benchmark, motivating decomposition of aggregate results by prediction target to reveal capability differences obscured by Overall summaries. Additional results are provided in Appendix~\ref{app:model-results}.

\begin{table*}[t]
\centering
\scriptsize

\setlength{\tabcolsep}{3pt}
\renewcommand{\arraystretch}{1.30}

\setlength{\aboverulesep}{0pt}
\setlength{\belowrulesep}{0pt}

\caption{\textbf{Multiple-choice performance under the blurred-image intervention.} Results are reported as \textbf{Acc.} /
\textcolor{iclrsubtext}{\textbf{F1}} (\%).
$\Delta$ Overall denotes the original-to-blurred score drop (pp)}
\label{tab:gaussian_blur_results}

\vspace{3pt}

\begin{tabularx}{\linewidth}{
>{\raggedright\arraybackslash}p{0.12\linewidth}
*{8}{Y}
}

\toprule

\rowcolor{iclrheader}
&
\textbf{Anatomy}
&
\multicolumn{2}{c}{
\cellcolor{iclrheader}\textbf{Finding}
}
&
\multicolumn{3}{c}{
\cellcolor{iclrheader}\textbf{Assessment \& Diagnosis}
}
&
\textbf{Overall}
&
\textbf{$\Delta$ Overall}
\\

\noalign{%
{\color{iclrheader}\hrule height \cmidrulewidth}%
\vskip-\cmidrulewidth
}

\cmidrule(lr){2-2}
\cmidrule(lr){3-4}
\cmidrule(lr){5-7}
\cmidrule(lr){8-8}
\cmidrule(lr){9-9}

\rowcolor{iclrheader}
\multirow{-2}{*}{\textbf{Model}}
&
\textbf{AA}
&
\textbf{FT}
&
\textbf{FA}
&
\textbf{AC}
&
\textbf{DX}
&
\textbf{PS}
&
\textbf{Acc/F1}
&
\textbf{Acc/F1}
\\

\midrule

Random Guess
& \clsresult{25.00}{20.64}
& \clsresult{12.50}{8.02}
& \clsresult{36.90}{30.95}
& \clsresult{19.76}{16.84}
& \clsresult{25.00}{22.39}
& \clsresult{41.67}{37.83}
& \clsresult{26.81}{22.78}
& \cellcolor{gray!10}{--}
\\

MedGemma-4B
& \clsresult{11.56}{11.58}
& \clsresult{8.24}{5.88}
& \clsresult{39.06}{27.99}
& \clsresult{1.83}{3.38}
& \clsresult{21.06}{16.73}
& \clsresult{40.44}{36.59}
& \clsresult{20.36}{17.02}
& \cellcolor{gray!10}{12.48 / 6.97}
\\

Lingshu-7B
& \clsresult{26.78}{19.60}
& \clsresult{16.48}{7.03}
& \clsresult{55.03}{27.96}
& \clsresult{32.85}{16.08}
& \clsresult{35.98}{18.74}
& \clsresult{61.92}{32.28}
& \clsresult{38.17}{20.28}
& \cellcolor{gray!10}{8.83 / 9.68}
\\

Hulu-Med-7B
& \clsresult{19.06}{16.62}
& \clsresult{5.71}{4.69}
& \clsresult{54.83}{28.07}
& \clsresult{31.07}{15.89}
& \clsresult{38.27}{22.01}
& \clsresult{52.23}{41.12}
& \clsresult{33.53}{21.40}
& \cellcolor{gray!10}{5.86 / 6.33}
\\

BreastGPT-8B
& \clsresult{36.66}{18.00}
& \clsresult{42.01}{11.48}
& \clsresult{58.54}{28.81}
& \clsresult{20.12}{11.37}
& \clsresult{30.70}{18.49}
& \clsresult{35.55}{21.97}
& \clsresult{37.26}{18.35}
& \cellcolor{gray!10}{17.97 / 22.61}
\\

\bottomrule

\end{tabularx}
\end{table*}
\paragraph{Annotation grounding provides a model-independent reference layer.} MedBenchAgent treats the annotations associated with each source image as the reference standard. This design places trust in clinician annotations rather than external model judgments, avoiding an additional source of uncertainty. To test this strategy, we replace every evaluation image with a Gaussian-blurred image while keeping all non-image benchmark components unchanged. Across four medical VLMs, Acc. decreases by 5.86–17.97 percentage points (pp) and F1 by 6.33–22.61 pp (Table~\ref{tab:gaussian_blur_results}). The consistent degradation indicates that benchmark performance remains sensitive to visual evidence from the corresponding image, despite references being derived directly from source annotations rather than model-based visual judgments.

\subsection{Cross-Domain Case Study}
\label{sec:cross-domain}

To assess the cross-domain portability of MedBenchAgent, we apply the complete construction workflow under a specialized medical setting using ISIC 2018~\citep{isic,HAM10000}, with a distinct dataset and annotation schema.

\setlength{\intextsep}{0pt}

\begin{wrapfigure}[9]{r}{0.60\linewidth}
    \centering
    \setlength{\abovecaptionskip}{2pt}
    \includegraphics[width=\linewidth]{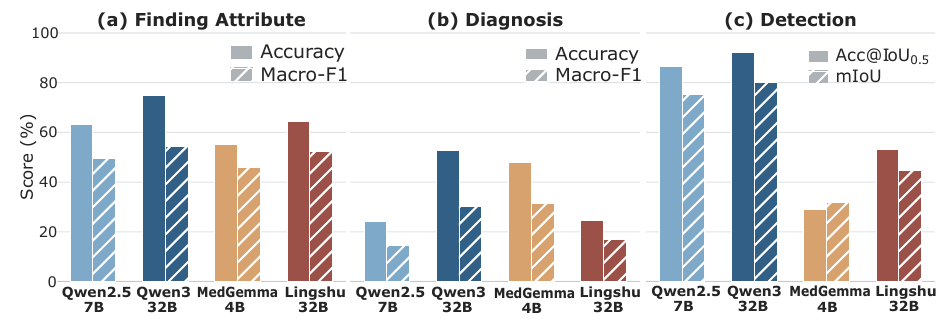}
    \caption{
        \textbf{Cross-domain instantiation on dermoscopy.} 
    }
    \label{fig:dermoscopy}
\end{wrapfigure}

The resulting task space and representative VLM results are summarized in Figure~\ref{fig:dermoscopy}. Representative VLMs show different relative strengths across prediction targets rather than a uniform performance pattern. This indicates that the same workflow can instantiate a domain-appropriate task structure under a distinct annotation schema These results provide cross-domain evidence for the portability of MedBenchAgent across annotation semantics and task definitions. Additional details are provided in Appendix~\ref{app:dermoscopy}.

\section{Conclusion}
We introduced MedBenchAgent, a multi-agent framework that treats medical VLM benchmark construction as constrained compilation and makes benchmark specification an explicit, auditable object of construction. Through the Benchmark Intermediate Representation (BIR), MedBenchAgent separates specification planning from downstream item instantiation, allowing construction decisions to be evaluated stage by stage rather than only through the final benchmark. Our analyses identify open-ended task-space specification as the primary source of construction uncertainty, while annotation-evidence mapping and item instantiation become substantially more reliable once benchmark semantics are fixed. Evaluation of the resulting benchmark further reveals task-, metric-, and imaging-dependent variation across VLMs that aggregate scores can obscure. Together with the cross-domain instantiation under a specialized medical setting, these results show that constrained, multi-agent benchmark construction can support both scalable medical VLM evaluation and systematic study of the benchmark-construction process itself.

\noindent{\textbf{Limitations and Future Directions.}}
While MedBenchAgent moves medical benchmark construction towards greater automation, limited human involvement remains at planning checkpoints; we will further reduce this reliance by introducing supervisory agents to verify intermediate specifications. More broadly, MedBenchAgent provides a basis for comparing controlled benchmark compositions and for constructing larger collections of disease-, anatomy-, and modality-specific benchmarks.

\bibliography{references}
\bibliographystyle{iclr2027_conference}

\appendix

\section{Constructed Breast-Imaging Benchmark}
\label{app:breast-benchmark}
\input{appendix/tables/tab_evidence_mapping}
\subsection{Raw Task Proposals and Evidence Mappings}
\label{app:raw-evidence-mappings}

Table~\ref{tab:app-raw-evidence-mappings} transcribes every task that TaskPlanningAgent marked \texttt{retain} in the initial \texttt{task\_plans/v1.json} output, before researcher confirmation. It shows the model's original task decomposition and its exact mapping from each proposed task to authorized annotation evidence. The count-affecting review decisions are reported separately in Appendix~\ref{app:task-space-results}.

\subsection{Audited Representative Items}
\label{app:representative-items}

The 20 entries in Figures~\ref{fig:breast1} and~\ref{fig:breast2} give one audited item for every atomic dataset--task pair instantiated in the realized breast-imaging benchmark, spanning BrEaST, BUS-BRA, and VinDr-Mammo. For each entry, we show the source dataset, task definition, representative image, generated query, and corresponding benchmark reference answer.

\begin{figure}[t]
    \centering
    \includegraphics[
        width=\linewidth,
        height=0.88\textheight,
        keepaspectratio
    ]{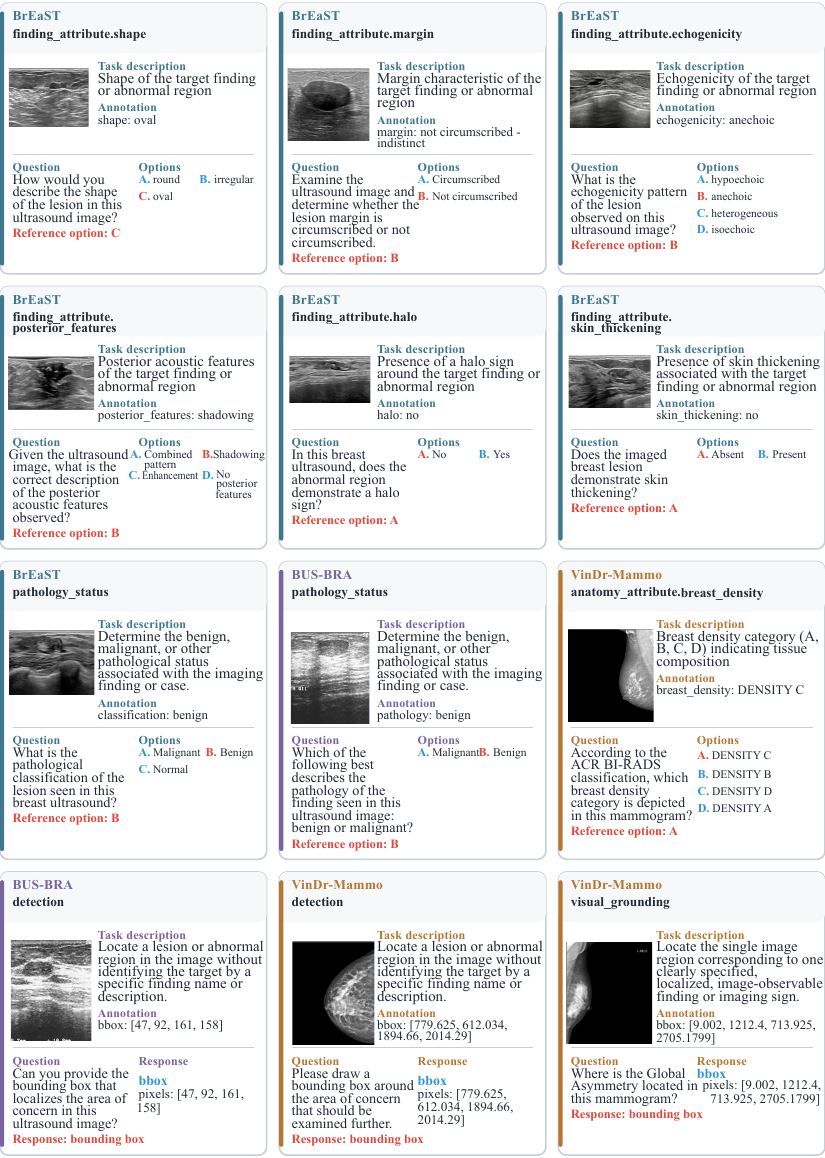}

    \setlength{\abovecaptionskip}{0pt}

    \caption{
        Representative audited items for 12 dataset--task mappings in the
        breast-imaging benchmark. Each card shows the source dataset and task,
        source annotation, representative image, generated query, and benchmark
        reference or spatial response.
    }
    \label{fig:breast1}
\end{figure}

\begin{figure}[t]
    \centering
    \includegraphics[
        width=\linewidth,
        height=0.88\textheight,
        keepaspectratio
    ]{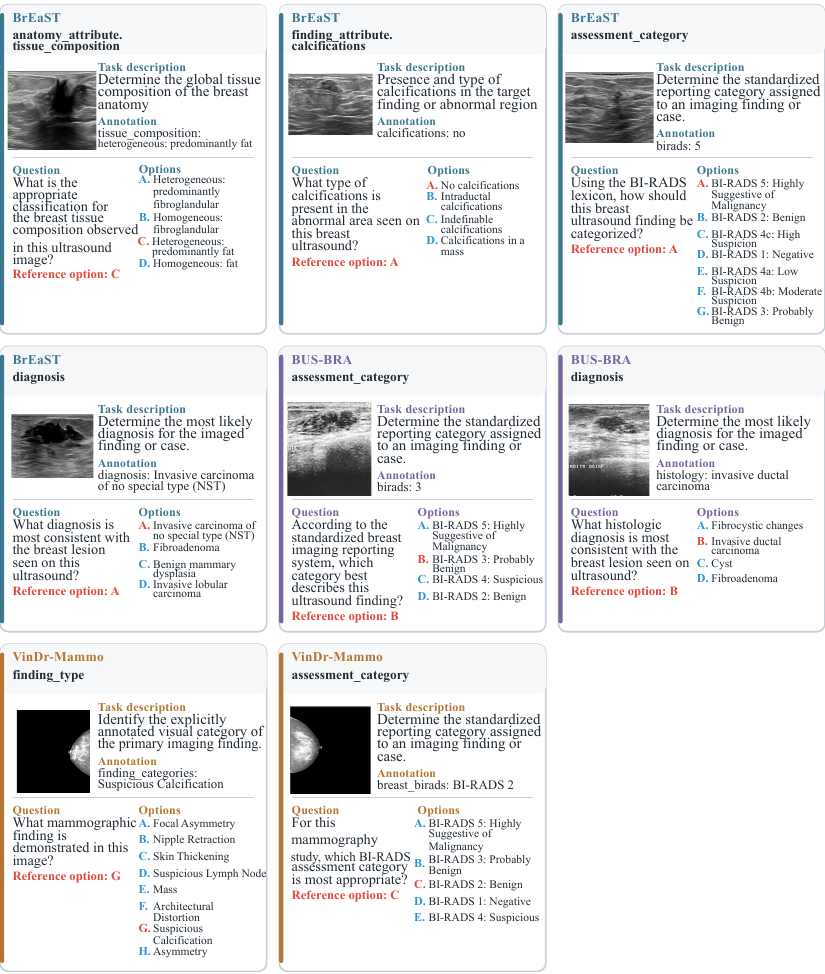}

    \setlength{\abovecaptionskip}{0pt}

    \caption{
        Continued from Figure~\ref{fig:breast1}. Representative samples from
        different dataset--task mappings, completing all 20 pairs.
    }
    \label{fig:breast2}
\end{figure}

\section{Construction Analysis and Benchmark Validation}
\label{app:construction}
\subsection{Reference Task Space and Task-Space Evaluation Protocol}
\label{app:baseline-setup}

\noindent\textbf{Baselines.}
We compare MedBenchAgent with three direct task-planning baselines to evaluate
BIR-mediated task-space compilation against direct task induction.
All methods receive the same confirmed evaluation scope $\mathcal{U}$ and the same dataset-specific profile derived from the normalized annotation schema $\Sigma_m$. MedBenchAgent compiles the initial task tree $\mathcal{T}_0$ within the BIR against the available annotation evidence through task retention and expansion. The direct baselines instead construct a complete set of prediction targets directly from $\mathcal{U}$ and $\Sigma_m$ in a single model call per dataset. This comparison therefore evaluates catalog-initialized BIR compilation against direct task induction under the same evaluation scope and dataset profile.
Direct $+\mathcal{T}_0$ augments the DeepSeek-V4-Flash direct baseline with the initial BIR task definitions as prompt reference.

For this comparison, both MedBenchAgent and the direct baselines are evaluated before human task-space confirmation. The MedBenchAgent result is the initial task-space proposal produced by TaskPlanningAgent, while each direct baseline is evaluated from its raw model output. All LLM baselines return a single JSON object through an OpenAI-compatible endpoint and are hosted locally on 1–2 NVIDIA B200 GPUs.

\noindent\textbf{Reference task-space design.}
\label{app:task-space-review}
We designed the breast-imaging reference task space $\mathcal{T}^{\star}$ from the perspective of medical VLM evaluation, with the goal of covering the major stages of imaging diagnosis across the breast cancer diagnostic workflow. Informed by discussions within the research team and with a clinical collaborator, we abstracted this workflow into eight evaluation phases: image recognition and validity, global tissue characterization, lesion detection and localization, finding characterization, diagnostic classification, standardized assessment, management and integrated reasoning, and report generation. These phases provide a clinically structured progression from basic image interpretation to increasingly integrated diagnostic decisions, while remaining expressed as model-evaluable visual prediction capabilities.

We then applied this staged framework using the task definitions and reference annotations available in the public breast-imaging datasets considered in this study. For each phase, we defined a prediction target only when the corresponding annotations provided a clear reference standard, and we represented distinct prediction targets as separate atomic tasks. The framework defines the intended clinical coverage, while each dataset determines which stages have annotation-supported targets. Stages without an eligible target remain in the framework, showing both the covered capabilities and the parts of the broader imaging workflow not supported by the available public annotations.

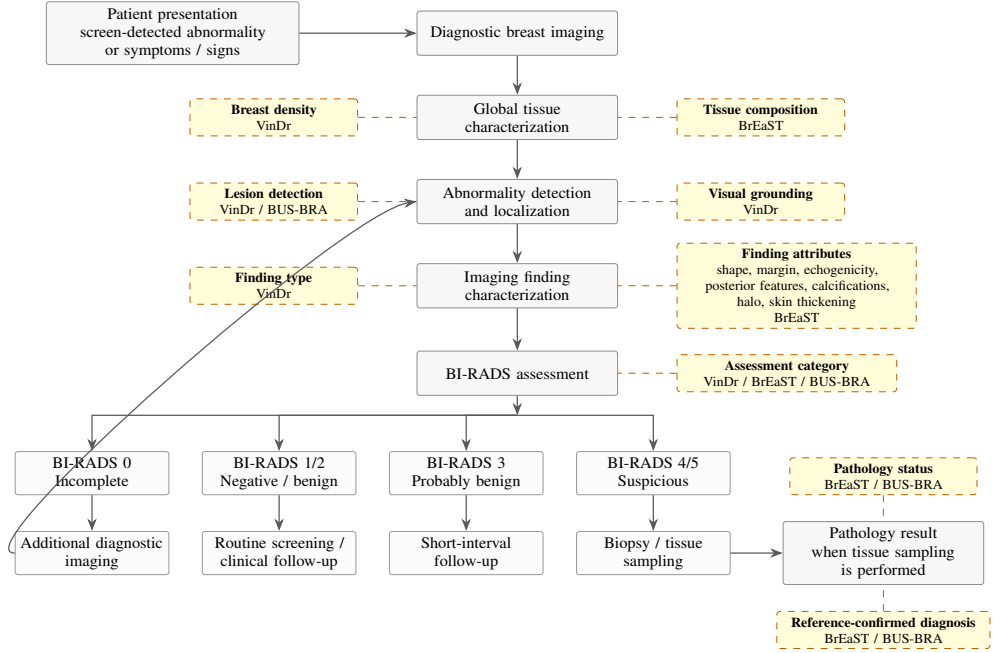
\begin{figure}[t]
\centering
\resizebox{0.98\linewidth}{!}{%

\begin{tikzpicture}[
    >=Stealth,
    font=\scriptsize,
    clinical/.style={
        draw=black!45,
        rounded corners=1.2pt,
        fill=gray!6,
        minimum height=6.8mm,
        text width=29mm,
        align=center,
        inner sep=2.6pt
    },
    branch/.style={
        draw=black!40,
        rounded corners=1.2pt,
        fill=gray!3,
        minimum height=6.5mm,
        text width=22mm,
        align=center,
        inner sep=2.3pt
    },
    task/.style={
        draw=orange!75!black,
        dashed,
        rounded corners=1.4pt,
        fill=yellow!18,
        font=\tiny,
        text width=24mm,
        align=center,
        inner sep=2.5pt
    },
    flow/.style={
        ->,
        draw=black!65,
        line width=0.55pt
    },
    tasklink/.style={
        draw=orange!75!black,
        dashed,
        line width=0.45pt
    }
]

\node[clinical] (input)
{Diagnostic breast imaging};

\node[clinical, text width=37mm, left=18mm of input, yshift=0mm] (entry)
{Patient presentation\\
screen-detected abnormality\\
or symptoms / signs};

\draw[flow] (entry.east) -- ++(4mm,0) |- (input.west);

\node[clinical, below=6mm of input] (global)
{Global tissue\\characterization};

\node[clinical, below=6mm of global] (localize)
{Abnormality detection\\and localization};

\node[clinical, below=6mm of localize] (finding)
{Imaging finding\\characterization};

\node[clinical, below=6.5mm of finding] (assessment)
{BI-RADS assessment};

\draw[flow] (input) -- (global);
\draw[flow] (global) -- (localize);
\draw[flow] (localize) -- (finding);
\draw[flow] (finding) -- (assessment);

\node[task, left=9mm of global] (t1)
{\textbf{Breast density}\\VinDr};

\node[task, right=9mm of global] (t2)
{\textbf{Tissue composition}\\BrEaST};

\node[task, left=9mm of localize] (t3)
{\textbf{Lesion detection}\\VinDr / BUS-BRA};

\node[task, right=9mm of localize] (t4)
{\textbf{Visual grounding}\\VinDr};

\node[task, left=9mm of finding] (t5)
{\textbf{Finding type}\\VinDr};

\node[task, right=9mm of finding, text width=35mm] (t6)
{\textbf{Finding attributes}\\
shape, margin, echogenicity,\\
posterior features, calcifications,\\
halo, skin thickening\\
BrEaST};

\node[task, right=9mm of assessment, text width=32mm] (t7)
{\textbf{Assessment category}\\
VinDr / BrEaST / BUS-BRA};

\draw[tasklink] (t1.east) -- (global.west);
\draw[tasklink] (t2.west) -- (global.east);

\draw[tasklink] (t3.east) -- (localize.west);
\draw[tasklink] (t4.west) -- (localize.east);

\draw[tasklink] (t5.east) -- (finding.west);
\draw[tasklink] (t6.west) -- (finding.east);

\draw[tasklink] (t7.west) -- (assessment.east);

\coordinate (split) at ($(assessment.south)+(0,-3mm)$);

\node[branch, below left=8.5mm and 38mm of assessment] (b0)
{BI-RADS 0\\Incomplete};

\node[branch, right=5mm of b0] (b12)
{BI-RADS 1/2\\Negative / benign};

\node[branch, right=5mm of b12] (b3)
{BI-RADS 3\\Probably benign};

\node[branch, right=5mm of b3] (b45)
{BI-RADS 4/5\\Suspicious};

\draw[flow] (assessment.south) -- (split);
\draw[flow] (split) -| (b0.north);
\draw[flow] (split) -| (b12.north);
\draw[flow] (split) -| (b3.north);
\draw[flow] (split) -| (b45.north);

\node[branch, below=5.5mm of b0] (addimg)
{Additional diagnostic\\imaging};

\node[branch, below=5.5mm of b12] (routine)
{Routine screening /\\clinical follow-up};

\node[branch, below=5.5mm of b3] (short)
{Short-interval\\follow-up};

\node[branch, below=5.5mm of b45] (biopsy)
{Biopsy / tissue\\sampling};

\draw[flow] (b0) -- (addimg);
\draw[flow] (b12) -- (routine);
\draw[flow] (b3) -- (short);
\draw[flow] (b45) -- (biopsy);

\draw[flow]
(addimg.west) .. controls +(-9mm,0) and +(-11mm,-1.5mm) .. (localize.west);

\node[clinical, right=8mm of biopsy, text width=29mm] (pathology)
{Pathology result\\
when tissue sampling\\
is performed};

\draw[flow] (biopsy.east) -- (pathology.west);

\node[task, above=4mm of pathology, text width=27mm] (t8)
{\textbf{Pathology status}\\BrEaST / BUS-BRA};

\node[task, below=4mm of pathology, text width=31mm] (t9)
{\textbf{Reference-confirmed diagnosis}\\BrEaST / BUS-BRA};

\draw[tasklink] (t8.south) -- (pathology.north);
\draw[tasklink] (t9.north) -- (pathology.south);

\end{tikzpicture}%
}
\caption{
Clinical breast-imaging workflow and its annotation-supported reference task space.
Tasks are attached to the corresponding stages of the imaging workflow.
}
\label{fig:clinical_task_space}
\end{figure}

Figure~\ref{fig:clinical_task_space} reports the resulting dataset-conditioned reference task space. Under the atomic matching protocol below, the table contains 21 reference prediction targets in total. A phase may contain multiple atomic targets when the available annotations support distinct prediction targets, whereas a categorical prediction target such as VinDr finding-type classification remains one task even when its answer space contains multiple finding categories.

\noindent\textbf{Independence from automated task-space construction.} The staged framework and the resulting $\mathcal{T}^{\star}$ were developed before the automated task-space experiments and were fixed prior to evaluating MedBenchAgent or any direct-construction baseline. $\mathcal{T}^{\star}$ was not provided to any evaluated construction method, and the reference task space was not added to, removed from, split, or merged in response to method outputs. It therefore serves only as the external reference for measuring agreement between automatically constructed task spaces and the independently designed breast-imaging evaluation space.

\noindent\textbf{Human involvement in task-space evaluation.} Development of the reference framework and adjudication of method outputs were separated from the automated construction procedures. The reference-space design involved benchmark-design discussions among dataset-familiar researchers with clinical input. For the Task-Space F1 analysis reported in this work, matching and adjudication between method proposals and the fixed reference task space were performed by a researcher using the predefined rules below.

\noindent\textbf{Prediction-target matching.} The unit of evaluation was a dataset-conditioned prediction target. An ordinary task counted once; a decomposable parent category did not count separately, whereas each distinct prediction target under that parent was evaluated as one task. Matching was based on whether the proposal and reference represented the same prediction target, regardless of task identifier, surface wording, evaluation metric, response format, or JSON nesting. A raw proposal was counted as a true positive (TP) when it matched one reference prediction target in $\mathcal{T}^\star$. A proposal without a corresponding reference task was counted as a false positive (FP), including unsupported, out-of-scope, shortcut-prone, or semantically redundant proposals. A reference task in $\mathcal{T}^{\star}$ without a semantically equivalent proposal was counted as a false negative (FN).

Matching was one-to-one. If multiple proposals represented the same reference prediction target, at most one was matched as a TP and each additional duplicate was counted as an FP. If a single proposal conflated multiple distinct prediction targets, it could match at most one reference task, with each remaining unmatched target counted as an FN. Differences that did not change the prediction target—including the evaluation metric, response format, bounding-box coordinate convention, surface wording, or exact annotation-field spelling—did not affect task-level matching.

\noindent\textbf{Evidence-Mapping Agreement.} For task proposals matched to $\mathcal{T}^{\star}$, we additionally evaluated whether the method identified the annotation fields required to support the corresponding reference construction. Evidence-Mapping Agreement (EMA) was computed only over matched TPs. For each matched task, agreement was assigned a value of one when the method-proposed supporting-field set was identical to the supporting-field set confirmed during review against the dataset annotation schema $\Sigma_m$, and zero otherwise. FPs and FNs were excluded from the EMA denominator.

\noindent\textbf{Metrics and aggregation.} For each method--dataset pair, we computed
\[
P=\frac{TP}{TP+FP},\qquad R=\frac{TP}{TP+FN},\qquad F_1=\frac{2TP}{2TP+FP+FN}.
\]
We additionally report the number of task-space discrepancies between a raw proposal and $\mathcal{T}^{\star}$, defined as $FP+FN$. Overall results micro-aggregate TP, FP, and FN across the three datasets before applying these equations. Thus, datasets contribute according to their numbers of reference and proposed prediction targets rather than receiving equal dataset weight.

\paragraph{Direct task-induction baseline.}

The direct baselines receive the same confirmed benchmark requirement and dataset profile as
MedBenchAgent, but do not receive the predefined task catalogue $T_0$. They induce the complete
task space in a single model call under the same output schema and evidence-admissibility rules,
including annotation-supported prediction targets, atomic answer axes, semantic-duplicate removal,
and explicit spatial-evidence requirements. The complete baseline prompts and serialization
instructions are provided with the released code.

\subsection{Task-Space Results and Discrepancy Analysis}
\label{app:task-space-results}

Tables~\ref{tab:app-revisions-a} and~\ref{tab:app-revisions-b} detail task-space discrepancies for MedBenchAgent and the three direct baselines. Their matched tasks have supporting-field mappings consistent with the reference, as reflected by the EMA results in Table~\ref{tab:task_design_review}.

\begin{table*}[t]
\centering
\scriptsize
\caption{
Task-space discrepancies affecting TP, FP, or FN counts for MedBenchAgent
and Direct DeepSeek. Retained true positives are omitted.
}
\label{tab:app-revisions-a}
\begin{tabularx}{\textwidth}{lll l X}
\toprule
Method & Dataset & Proposal or confirmed task & Label & Review rationale \\
\midrule
MedBenchAgent & BUS-BRA & finding type & FP & Duplicated diagnosis on the histology answer axis. \\
MedBenchAgent & BUS-BRA & visual grounding & FP & Did not define a capability distinct from generic lesion detection. \\
MedBenchAgent & VinDr & view identification & FP & Burned-in view markers exposed the target as a shortcut. \\
MedBenchAgent & VinDr & finding-level assessment & FN & Finding-level BI-RADS supported a distinct, initially omitted answer axis. \\
\midrule
Direct DeepSeek & BrEaST & clinical signs & FP & Nonvisual clinical metadata. \\
Direct DeepSeek & BrEaST & patient symptoms & FP & Nonvisual patient information. \\
Direct DeepSeek & BrEaST & verification method & FP & Nonvisual examination/pathology metadata. \\
Direct DeepSeek & BrEaST & tumor localization & FP & Treated a mask as bounding-box evidence. \\
Direct DeepSeek & BrEaST & diagnosis & FN & The explicit diagnosis field was not proposed as an independent task. \\
Direct DeepSeek & BUS-BRA & side identification & FP & Laterality metadata was outside the intended visual scope. \\
Direct DeepSeek & BUS-BRA & device identification & FP & Acquisition-device metadata was outside the clinical visual scope. \\
Direct DeepSeek & VinDr & laterality classification & FP & Metadata/marker shortcut rather than the intended image semantics. \\
Direct DeepSeek & VinDr & view-position classification & FP & Burned-in view marker leaked the target. \\
Direct DeepSeek & VinDr & visual grounding & FN & Finding category was not paired with its box. \\
\bottomrule
\end{tabularx}
\end{table*}

\begin{table*}[t]
\centering
\scriptsize
\caption{
Task-space discrepancies affecting TP, FP, or FN counts for Direct Qwen
and Direct MedGemma. Retained true positives are omitted.
}
\label{tab:app-revisions-b}
\begin{tabularx}{\textwidth}{lll l X}
\toprule
Method & Dataset & Proposal or confirmed task & Label & Review rationale \\
\midrule
Direct Qwen & BrEaST & palpability assessment & FP & Clinical palpation is not image-answerable. \\
Direct Qwen & BrEaST & interpretation classification & FP & A source conclusion is not an independent visual capability. \\
Direct Qwen & BrEaST & verification method & FP & Nonvisual confirmation metadata. \\
Direct Qwen & BrEaST & patient symptom status & FP & Nonvisual patient information. \\
Direct Qwen & BUS-BRA & lateralization & FP & Laterality metadata was outside the intended scope. \\
Direct Qwen & BUS-BRA & device identification & FP & Acquisition-device metadata was outside the intended scope. \\
Direct Qwen & VinDr & laterality classification & FP & Metadata/marker shortcut. \\
Direct Qwen & VinDr & view-position classification & FP & Burned-in view marker leaked the target. \\
Direct Qwen & VinDr & breast density & FN & A varying breast-density field supported an omitted global answer axis. \\
Direct Qwen & VinDr & visual grounding & FN & Finding category was not paired with its box. \\
\midrule
Direct MedGemma & BrEaST & anatomy classification & FP & Anatomy was constant and could not form a valid classification task. \\
Direct MedGemma & BrEaST & modality classification & FP & Modality was constant. \\
Direct MedGemma & BrEaST & interpretation classification & FP & Source interpretation was not an independent visual capability. \\
Direct MedGemma & BrEaST & age classification & FP & Patient age was nonvisual metadata. \\
Direct MedGemma & BrEaST & patient symptoms & FP & Patient symptoms were nonvisual. \\
Direct MedGemma & BrEaST & signs classification & FP & Clinical-sign metadata was outside the target visual scope. \\
Direct MedGemma & BrEaST & verification classification & FP & Nonvisual confirmation metadata. \\
Direct MedGemma & VinDr & laterality classification & FP & Metadata/marker shortcut. \\
Direct MedGemma & VinDr & view-position classification & FP & Burned-in view marker leaked the target. \\
Direct MedGemma & VinDr & visual grounding & FN & Finding category was not paired with its box. \\
\bottomrule
\end{tabularx}
\end{table*}

The discrepancy patterns also differ qualitatively across construction strategies. 
MedBenchAgent's residual errors are concentrated in semantic boundary cases, including a redundant BUS-BRA finding-type proposal, a visual-grounding proposal that did not define a capability distinct from generic detection, a shortcut-prone VinDr view-identification task, and the omitted finding-level assessment target. 
In contrast, the direct baselines more frequently introduce nonvisual metadata, acquisition or marker shortcuts, invalid spatial evidence, or omit otherwise supported visual prediction targets. 
This distinction helps explain why Evidence-Mapping Agreement is perfect for matched tasks even when task-space precision differs: once the intended prediction target is identified, annotation-field selection is comparatively stable, whereas deciding which targets should constitute the benchmark remains more difficult.
\subsection{Human Item Audit and Residual Errors}
\label{app:human-audit}

\noindent\textbf{Sampling and review setting.}
The audit population comprised all 16,248 finalized items accepted by the
automatic screening and audit procedures in the completed BrEaST, BUS-BRA,
and VinDr-Mammo runs. We globally shuffled this combined population and sampled 1,000 items uniformly without replacement.

We reviewed each sampled item using the predefined criteria below. The review materials comprised the dataset and task identifiers, response format, question, options, benchmark reference, and complete normalized annotation. The audit evaluated construction consistency against the source annotations and locked task specification.

\noindent\textbf{Audit criteria.}
Four binary criteria were applied.
\emph{Reference groundedness} required the benchmark reference to follow directly from the authorized annotation evidence.
\emph{Task--query alignment} required the question to express the locked prediction target.
\emph{Unambiguity} required the question to admit one intended answer under
the supplied evidence.
\emph{Evaluation integrity} required the absence of answer leakage, overlapping
response options, question--reference mismatch, response-format mismatch, and
unsupported additions.
An item was considered valid only when all four criteria passed:
\[
\mathrm{ValidItem}_i
=
\mathbf{1}\{G_i \land A_i \land U_i \land I_i\},
\]
where $G_i$, $A_i$, $U_i$, and $I_i$ denote reference groundedness,
task--query alignment, unambiguity, and evaluation integrity, respectively.

\noindent\textbf{Residual errors.}
\label{app:residual-errors}
Six of the 1,000 sampled items failed at least one audit criterion. Four VinDr-Mammo
finding-type items exhibited the same template-level task--query mismatch: the generated question
asked for a BI-RADS assessment category while the locked answer space contained finding categories.
Two BrEaST diagnosis items introduced unsupported contextual details and therefore failed evaluation
integrity. Importantly, the fixed annotation-derived reference answers remained supported by the
authorized evidence in all six cases. Across the complete audit, there were four task--query-alignment
failures and six evaluation-integrity failures, with no reference-groundedness or unambiguity failures.

\section{Cross-Domain Portability Study: Dermoscopy}
\label{app:dermoscopy}
\subsection{Dataset, Task Space, and Realized Allocation}
\label{app:dermoscopy-setup}

The cross-domain run used the ISIC 2018 challenge data from its three official task sources: lesion segmentation (Task~1), dermoscopic-attribute detection (Task~2), and disease classification (Task~3). Tasks~1 and~2 share 1,000 held-out dermoscopic images, while Task~3 contributes 1,512 held-out diagnosis images. After deduplicating the shared Task~1/2 identifiers, the normalized corpus contains 2,512 unique images. Diagnosis labels are available for the Task~3 subset, whereas lesion masks and five dermoscopic-attribute masks are available for the Task~1/2 subset. Because these annotations are defined on different image subsets, no single task is eligible for all 2,512 images. We apply the workflow to the ISIC annotation schema and evaluate the resulting task-to-evidence mappings, item consistency, and model performance.

\noindent\textbf{Task space and realized allocation.}
\label{app:dermoscopy-task-space}
The requested 5,000-item allocation yielded 4,179 realized items because of source-data eligibility limits. All eligible diagnosis (1,512) and lesion-localization (1,000) cases were retained, with the remaining 1,667 items allocated across the five dermoscopic-attribute tasks. We checked the initial task plan against the documented ISIC annotation schema and label semantics. All seven proposed tasks and their supporting-field mappings were retained without task-space revision.

\begin{table*}[t]
\centering
\scriptsize
\caption{ISIC 2018 tasks, supporting annotation fields, and realized item counts.}
\label{tab:app-dermoscopy-tasks}
\begin{tabular}{lllrrr}
\toprule
Task & Supporting field & Resp. & Metric & Planned & Realized \\
\midrule
Diagnosis
& \path{image.annotation.diagnosis}
& letter
& Macro-F1
& 1,667
& 1,512 \\

Detection
& \makecell[l]{\path{image.annotation.findings[*]}\\\path{.bbox[*]}}
& bbox
& mIoU
& 1,666
& 1,000 \\

Globules presence
& \makecell[l]{\path{image.annotation.attribute_presence}\\\path{.globules}}
& letter
& Accuracy
& 334
& 334 \\

Milia-like cyst presence
& \makecell[l]{\path{image.annotation.attribute_presence}\\\path{.milia_like_cyst}}
& letter
& Macro-F1
& 334
& 334 \\

Negative network presence
& \makecell[l]{\path{image.annotation.attribute_presence}\\\path{.negative_network}}
& letter
& Macro-F1
& 333
& 333 \\

Pigment network presence
& \makecell[l]{\path{image.annotation.attribute_presence}\\\path{.pigment_network}}
& letter
& Accuracy
& 333
& 333 \\

Streaks presence
& \makecell[l]{\path{image.annotation.attribute_presence}\\\path{.streaks}}
& letter
& Macro-F1
& 333
& 333 \\

\midrule
\multicolumn{4}{r}{Total} & 5,000 & 4,179 \\
\bottomrule
\end{tabular}
\end{table*}

\subsection{Item Consistency and Representative Examples}
\label{app:dermoscopy-validation}

\paragraph{Item consistency.}
We additionally performed a consistency check on 200 finalized ISIC items using the same
annotation- and contract-based criteria as the breast-imaging audit. All sampled items passed checks
of reference--evidence consistency, task--query consistency, and evaluation-format integrity against
the normalized annotations and locked task contracts.

\begin{figure}[t]
    \centering
    \includegraphics[width=\linewidth]{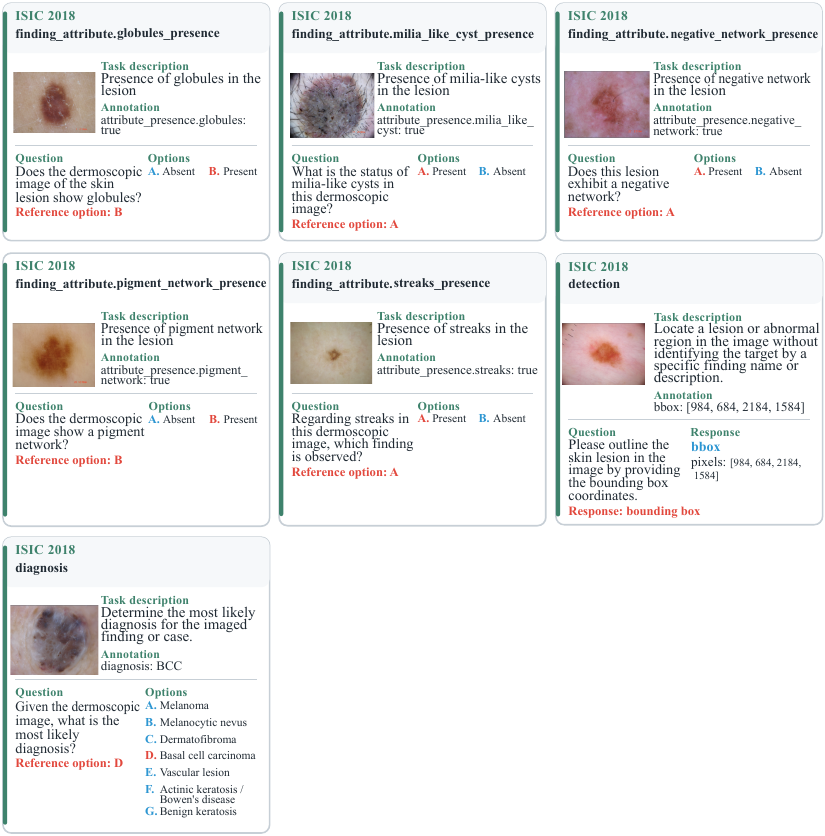}
    \setlength{\abovecaptionskip}{0pt}
    \caption{
        Representative audited ISIC 2018 items for all seven dermoscopy tasks.}
    \label{fig:dermoscopy_items}
\end{figure}

Figure~\ref{fig:dermoscopy_items} provides one audited representative item for
each of the seven dermoscopy tasks using the same presentation format as
the breast-imaging examples.

\subsection{Model Evaluation}
\label{app:dermoscopy-model-results}

Table~\ref{tab:app-isic-model-results} reports performance across diagnosis, dermoscopic attributes, and localization. Different checkpoints lead on different task metrics, providing complementary evaluation profiles. Together with the task-space review and item audit, these results demonstrate application of MedBenchAgent to the ISIC annotation schema.

\begin{table*}[t]
\centering
\scriptsize

\setlength{\tabcolsep}{3.6pt}
\renewcommand{\arraystretch}{1.30}

\setlength{\aboverulesep}{0pt}
\setlength{\belowrulesep}{0pt}

\caption{
Evaluation performance on ISIC 2018. Choice tasks report Accuracy /
Macro-F1 (\%), while detection reports mIoU / Acc@IoU0.5 (\%). Best results for
each metric are shown in bold.
}
\label{tab:app-isic-model-results}

\vspace{3pt}

\begin{tabularx}{\linewidth}{
>{\raggedright\arraybackslash}p{0.14\linewidth}
*{7}{Y}
}
\toprule

\rowcolor{iclrheader}
&
&
\multicolumn{5}{c}{\cellcolor{iclrheader}\textbf{Finding Attribute}}
&
\\

\noalign{%
    {\color{iclrheader}\hrule height \cmidrulewidth}%
    \vskip-\cmidrulewidth
}

\cmidrule(lr){3-7}

\rowcolor{iclrheader}
\multirow{-2}{*}{\textbf{Model}}
&
\multirow{-2}{*}{\textbf{Diagnosis}}
&
\textbf{Globules}
&
\textbf{Milia-like}
&
\textbf{Neg. net.}
&
\textbf{Pigment net.}
&
\textbf{Streaks}
&
\multirow{-2}{*}{\textbf{Detection}}
\\

\midrule

\rowcolor{gray!10}
\multicolumn{8}{l}{\textbf{\textit{General-purpose VLMs}}}
\\

InternVL3-8B
& \clsresult{20.7}{15.3}
& \clsresult{31.4}{31.1}
& \clsresult{70.7}{56.6}
& \clsresult{45.3}{35.0}
& \clsresult{72.4}{63.4}
& \clsresult{39.3}{35.7}
& \spresult{50.5}{52.2}
\\

Qwen2.5-VL-7B
& \clsresult{24.2}{14.6}
& \clsresult{57.8}{53.7}
& \clsresult{80.8}{55.5}
& \clsresult{83.8}{47.4}
& \clsresult{59.5}{59.3}
& \clsresult{33.6}{30.7}
& \spresult{75.1}{86.6}
\\

Qwen3-VL-8B-I
& \clsresult{53.2}{22.8}
& \clsresult{28.4}{27.5}
& \clsresult{80.2}{\textbf{61.8}}
& \clsresult{80.2}{47.3}
& \clsresult{59.5}{58.4}
& \clsresult{58.0}{44.7}
& \spresult{\textbf{80.3}}{91.4}
\\

Qwen3-VL-8B-T$^\dagger$
& \clsresult{30.9}{18.6}
& \clsresult{60.5}{52.9}
& \clsresult{83.2}{61.1}
& \clsresult{85.0}{48.2}
& \clsresult{57.1}{56.0}
& \clsresult{82.9}{47.0}
& \spresult{71.2}{83.6}
\\

Qwen2.5-VL-32B
& \clsresult{29.8}{20.8}
& \clsresult{25.1}{23.6}
& \clsresult{46.7}{40.2}
& \clsresult{28.5}{24.3}
& \clsresult{72.4}{64.5}
& \clsresult{15.9}{15.9}
& \spresult{76.8}{87.6}
\\

Qwen3-VL-32B
& \clsresult{52.8}{30.3}
& \clsresult{70.1}{58.2}
& \clsresult{84.4}{55.1}
& \clsresult{74.2}{45.8}
& \clsresult{58.0}{57.3}
& \clsresult{\textbf{88.0}}{\textbf{55.1}}
& \spresult{80.2}{\textbf{92.2}}
\\

\midrule

\rowcolor{gray!10}
\multicolumn{8}{l}{\textbf{\textit{Medical VLMs}}}
\\

MedGemma-4B
& \clsresult{47.8}{31.5}
& \clsresult{25.7}{24.0}
& \clsresult{66.8}{52.5}
& \clsresult{45.3}{33.5}
& \clsresult{\textbf{79.0}}{\textbf{71.7}}
& \clsresult{59.5}{47.7}
& \spresult{31.8}{29.1}
\\

Lingshu-7B
& \clsresult{10.6}{7.0}
& \clsresult{68.3}{57.2}
& \clsresult{82.9}{53.9}
& \clsresult{46.8}{36.2}
& \clsresult{76.9}{71.4}
& \clsresult{46.5}{38.4}
& \spresult{11.7}{6.8}
\\

Hulu-Med-7B
& \clsresult{\textbf{56.3}}{\textbf{34.4}}
& \clsresult{\textbf{75.4}}{47.4}
& \clsresult{83.8}{59.0}
& \clsresult{\textbf{95.2}}{\textbf{48.8}}
& \clsresult{40.8}{38.6}
& \clsresult{76.6}{48.9}
& \spresult{0.7}{0.2}
\\

QoQ-Med-7B$^\dagger$
& \clsresult{49.5}{21.4}
& \clsresult{69.2}{48.2}
& \clsresult{\textbf{84.7}}{55.6}
& \clsresult{90.7}{47.6}
& \clsresult{41.1}{39.2}
& \clsresult{75.7}{46.8}
& \spresult{25.8}{11.3}
\\

Lingshu-32B
& \clsresult{24.6}{16.8}
& \clsresult{71.9}{\textbf{60.3}}
& \clsresult{75.7}{53.5}
& \clsresult{44.1}{33.9}
& \clsresult{72.4}{69.2}
& \clsresult{57.4}{44.8}
& \spresult{44.5}{53.3}
\\

\bottomrule

\end{tabularx}
\end{table*}

Table~\ref{tab:app-isic-model-results} shows that the cross-domain benchmark preserves substantial variation across task families and model checkpoints. 
No single checkpoint exhibits uniformly strong performance across diagnosis, dermoscopic attributes, and lesion localization, and the relative behavior of general-purpose and medical VLMs varies across these task types. 
In particular, categorical performance does not reliably predict spatial-localization performance. 
These results are consistent with the breast-imaging evaluation in showing that aggregate model quality cannot be inferred from one task family alone, while providing evidence that the same construction workflow can be instantiated under a different annotation schema and clinical vocabulary.





\section{Additional Model-Evaluation Results}
\label{app:model-results}
\subsection{Evaluation Protocol}
\label{app:evaluation-protocol}
For reporting, fine-grained question types in Table~\ref{tab:main_results} are first scored separately and macro-averaged within their parent task where applicable, while dataset-specific results for the same task family are then macro-averaged with equal weight, and the reported Overall score macro-averages the resulting task-family scores. 

In this section, we report the complete per-dataset model-evaluation results and detailed evaluation protocol. Tables~\ref{tab:app-choice-breast}, \ref{tab:app-choice-busbra}, and~\ref{tab:app-choice-vindr} use the same two-level template to report dataset-level task results for BrEaST, BUS-BRA, and VinDr-Mammo. The abbreviations are AA (anatomy attribute), FA (finding attribute), FT (finding type), AC (assessment category), DX (diagnosis), and PS (pathology status). Table~\ref{tab:app-choice-breast-attributes} separately expands BrEaST FA into its seven question types. Table~\ref{tab:app-spatial-results} reports mIoU and $\mathrm{Acc@IoU}_{0.5}$ for each dataset--task pair. Scores are computed over all evaluated items, assigning zero IoU when no valid bounding box is returned.

\paragraph{Inputs and response parsing.}
Each model received the image, question, and lettered options when applicable, without the reference answer. Model-specific chat templates were applied with deterministic decoding. Multiple-choice predictions were parsed as a single valid option letter and evaluated using Accuracy and Macro-F1 over semantic answer classes. Invalid responses remained incorrect. Spatial predictions were parsed as a single four-number bounding box, accepting the supported \texttt{bbox} or \texttt{bbox\_2d} fields, bare lists or tuples, and JSON-fenced outputs. Missing, malformed, nonfinite, degenerate, out-of-range, or multiple distinct boxes were invalid and received zero IoU.

\paragraph{Coordinate conversion and scoring.}
We strictly follow the protocols of all models that claim to have spatial evaluation capabilities. Predictions were converted from the configured model-specific coordinate representation to absolute original-image coordinates before scoring. Qwen3-VL, InternVL3, and BreastGPT used a fixed $[0,1000]$ coordinate space; Qwen2.5-VL, Lingshu, and QoQ-Med used coordinates in the resized model input; and Hulu-Med used original-image coordinates. MedGemma used normalized $[0,1]$ coordinates in $[y_1,x_1,y_2,x_2]$ order, which were reordered and rescaled. The effective coordinate size was recorded with each prediction. Spatial metrics were computed over all evaluated items, including invalid outputs. Accordingly, mIoU and $\mathrm{Acc@IoU}_{0.5}$ measure end-to-end bounding-box prediction under the specified response protocol.

\subsection{Pairwise Model-Ordering Analysis}
\label{app:ranking-analysis}

The summary statistics of model-ordering reversals are reported in Section~\ref{sec:mainresults}. Here, we provide the detailed pairwise model-ordering analysis and describe how these statistics are computed. For two models $i$ and $j$, an ordering reversal between scores $s$ and $s'$ occurs when
\[
(s_i-s_j)(s'_i-s'_j)<0.
\]
With 12 evaluated VLMs, there are $\binom{12}{2}=66$ unordered model pairs. Ties are not counted as reversals.

\paragraph{Overall versus task-level evaluation.}
We compare the ordering induced by Overall Acc. in Table~\ref{tab:main_results} with the Acc. ordering for each task category. The numbers of reversed pairs are 19 (AA), 22 (FT), 18 (FA), 24 (AC), 12 (DX), and 22 (PS), corresponding to 18.2--36.4\% of all model pairs.

\paragraph{Acc. versus F1.}
Within each task category, we compare model ordering under Acc. and F1. The numbers of reversed pairs are 13 (AA), 11 (FT), 5 (FA), 15 (AC), 15 (DX), and 4 (PS), corresponding to 6.1--22.7\%.

\paragraph{Choice-based versus spatial evaluation.}
We compare Overall Acc. with each of the six spatial measures in Table~\ref{tab:spatial_results}. The numbers of reversed model pairs are 33 and 32 for VinDr-Mammo detection, 34 and 32 for VinDr-Mammo visual grounding, and 29 and 28 for BUS-BRA detection, for mIoU and $\mathrm{Acc@IoU}_{0.5}$, respectively. Thus, 28--34 of the 66 model pairs (42.4--51.5\%) reverse their ordering across the spatial measures.

\subsection{Additional Analysis of Model Behavior}
\label{app:model-analysis}
\begin{table*}[t]
\centering
\scriptsize

\setlength{\tabcolsep}{3.6pt}
\renewcommand{\arraystretch}{1.30}

\setlength{\aboverulesep}{0pt}
\setlength{\belowrulesep}{0pt}

\caption{\textbf{Multiple-choice evaluation on BrEaST by task type.} Results are reported as \textbf{Accuracy} / \textcolor{iclrsubtext}{\textbf{Macro-F1}} (\%). AA contains 200 tissue-composition items; FA equally averages its seven question types; AC and PS each contain 256 items; and DX contains 121 items. $\dagger$ denotes thinking variants.}
\label{tab:app-choice-breast}

\vspace{3pt}

\begin{tabularx}{0.70\linewidth}{
>{\raggedright\arraybackslash}p{0.18\linewidth}
*{5}{Y}
}

\toprule

\rowcolor{iclrheader}
&
\textbf{Anatomy}
&
\textbf{Finding}
&
\multicolumn{3}{c}{\cellcolor{iclrheader}\textbf{Assessment \& Diagnosis}}
\\

\noalign{%
    {\color{iclrheader}\hrule height \cmidrulewidth}%
    \vskip-\cmidrulewidth
}

\cmidrule(lr){2-2}
\cmidrule(lr){3-3}
\cmidrule(lr){4-6}

\rowcolor{iclrheader}
\multirow{-2}{*}{\textbf{Model}}
& \textbf{AA}
& \textbf{FA}
& \textbf{AC}
& \textbf{DX}
& \textbf{PS}
\\

\midrule

\rowcolor{gray!10}
\multicolumn{6}{l}{\textbf{\textit{General-purpose VLMs}}}
\\
InternVL3-8B & \clsresult{27.5}{10.8} & \clsresult{42.3}{30.9} & \clsresult{18.8}{10.1} & \clsresult{28.9}{16.3} & \clsresult{50.8}{33.4} \\
Qwen2.5-VL-7B & \clsresult{33.0}{19.5} & \clsresult{41.5}{29.3} & \clsresult{16.4}{10.5} & \clsresult{34.7}{21.3} & \clsresult{54.7}{36.4} \\
Qwen3-VL-8B-I& \clsresult{27.5}{10.9} & \clsresult{29.8}{22.6} & \clsresult{20.7}{16.7} & \clsresult{38.8}{26.9} & \clsresult{60.9}{40.3} \\
Qwen3-VL-8B-T$^\dagger$ & \clsresult{28.0}{13.4} & \clsresult{56.9}{35.1} & \clsresult{21.1}{19.3} & \clsresult{38.0}{24.5} & \clsresult{49.2}{32.7} \\
Qwen2.5-VL-32B & \clsresult{25.0}{15.2} & \clsresult{46.2}{29.6} & \clsresult{19.1}{13.2} & \clsresult{48.8}{27.3} & \clsresult{47.3}{31.6} \\
Qwen3-VL-32B & \clsresult{17.5}{12.9} & \clsresult{37.1}{28.4} & \clsresult{20.7}{20.8} & \clsresult{35.5}{27.7} & \clsresult{62.9}{53.0} \\

\midrule
\rowcolor{gray!10}
\multicolumn{6}{l}{\textbf{\textit{Medical VLMs}}}
\\
MedGemma-4B & \clsresult{26.5}{16.0} & \clsresult{47.5}{33.9} & \clsresult{19.5}{13.4} & \clsresult{44.6}{32.4} & \clsresult{51.6}{36.9} \\
Lingshu-7B & \clsresult{\textbf{33.5}}{\textbf{24.0}} & \clsresult{56.8}{38.7} & \clsresult{24.6}{25.0} & \clsresult{46.3}{18.8} & \clsresult{\textbf{69.1}}{40.4} \\
Hulu-Med-7B & \clsresult{27.0}{11.3} & \clsresult{51.1}{35.5} & \clsresult{20.7}{16.0} & \clsresult{44.6}{24.4} & \clsresult{61.7}{41.3} \\
QoQ-Med-7B$^\dagger$ & \clsresult{30.0}{20.7} & \clsresult{66.1}{37.8} & \clsresult{20.7}{13.7} & \clsresult{43.0}{23.5} & \clsresult{48.8}{31.5} \\
BreastGPT-8B & \clsresult{21.5}{19.9} & \clsresult{\textbf{75.2}}{\textbf{52.5}} & \clsresult{\textbf{36.3}}{\textbf{36.1}} & \clsresult{\textbf{60.3}}{\textbf{39.7}} & \clsresult{68.4}{\textbf{55.5}} \\
Lingshu-32B & \clsresult{31.5}{22.2} & \clsresult{57.5}{43.0} & \clsresult{23.4}{18.7} & \clsresult{47.1}{20.3} & \clsresult{67.2}{42.3} \\

\bottomrule
\end{tabularx}
\end{table*}
\begin{table*}[!htbp]
\centering
\scriptsize

\setlength{\tabcolsep}{3.6pt}
\renewcommand{\arraystretch}{1.30}

\setlength{\aboverulesep}{0pt}
\setlength{\belowrulesep}{0pt}

\caption{\textbf{Multiple-choice evaluation on BrEaST: finding-attribute question types.} Results are reported as \textbf{Accuracy} / \textcolor{iclrsubtext}{\textbf{Macro-F1}} (\%). Best results for each metric are shown in bold.}
\label{tab:app-choice-breast-attributes}

\vspace{3pt}

\begin{tabularx}{0.85\linewidth}{
>{\raggedright\arraybackslash}p{0.18\linewidth}
*{7}{Y}
}

\toprule

\rowcolor{iclrheader}
&
\multicolumn{7}{c}{\cellcolor{iclrheader}\textbf{Finding Attribute}}
\\

\noalign{%
    {\color{iclrheader}\hrule height \cmidrulewidth}%
    \vskip-\cmidrulewidth
}

\cmidrule(lr){2-8}

\rowcolor{iclrheader}
\multirow{-2}{*}{\textbf{Model}}
& \textbf{Shape}
& \textbf{Margin}
& \textbf{Echo.}
& \textbf{Posterior}
& \textbf{Calc.}
& \textbf{Halo}
& \textbf{Skin}
\\

\midrule

\rowcolor{gray!10}
\multicolumn{8}{l}{\textbf{\textit{General-purpose VLMs}}}
\\
InternVL3-8B & \clsresult{54.5}{30.9} & \clsresult{53.2}{53.0} & \clsresult{46.3}{33.1} & \clsresult{21.8}{15.6} & \clsresult{45.2}{23.0} & \clsresult{26.0}{24.4} & \clsresult{48.8}{35.9} \\
Qwen2.5-VL-7B & \clsresult{55.4}{26.8} & \clsresult{58.3}{47.4} & \clsresult{40.3}{28.3} & \clsresult{19.4}{8.1} & \clsresult{27.8}{14.7} & \clsresult{47.6}{44.6} & \clsresult{41.4}{35.1} \\
Qwen3-VL-8B-I& \clsresult{34.3}{28.6} & \clsresult{47.2}{42.2} & \clsresult{25.5}{16.2} & \clsresult{37.3}{24.8} & \clsresult{50.4}{24.0} & \clsresult{11.4}{16.7} & \clsresult{2.7}{5.6} \\
Qwen3-VL-8B-T$^\dagger$ & \clsresult{56.6}{34.4} & \clsresult{58.3}{45.8} & \clsresult{61.5}{28.4} & \clsresult{22.2}{12.4} & \clsresult{59.9}{30.4} & \clsresult{63.4}{46.6} & \clsresult{76.6}{47.8} \\
Qwen2.5-VL-32B & \clsresult{57.4}{32.3} & \clsresult{55.2}{45.3} & \clsresult{61.0}{35.1} & \clsresult{22.2}{12.4} & \clsresult{71.0}{28.6} & \clsresult{47.2}{43.8} & \clsresult{9.4}{9.4} \\
Qwen3-VL-32B & \clsresult{58.7}{39.9} & \clsresult{32.9}{40.3} & \clsresult{45.0}{34.7} & \clsresult{23.8}{17.9} & \clsresult{52.0}{24.7} & \clsresult{15.9}{17.3} & \clsresult{31.2}{24.2} \\

\midrule
\rowcolor{gray!10}
\multicolumn{8}{l}{\textbf{\textit{Medical VLMs}}}
\\
MedGemma-4B & \clsresult{52.9}{40.8} & \clsresult{54.8}{52.8} & \clsresult{39.8}{23.3} & \clsresult{52.8}{24.6} & \clsresult{29.0}{14.2} & \clsresult{42.3}{39.6} & \clsresult{61.3}{42.3} \\
Lingshu-7B & \clsresult{58.3}{45.6} & \clsresult{63.9}{60.2} & \clsresult{63.6}{32.6} & \clsresult{13.9}{9.4} & \clsresult{44.0}{18.8} & \clsresult{65.4}{48.5} & \clsresult{88.7}{55.5} \\
Hulu-Med-7B & \clsresult{58.3}{34.1} & \clsresult{59.5}{50.0} & \clsresult{58.9}{38.6} & \clsresult{25.8}{19.1} & \clsresult{17.9}{13.3} & \clsresult{64.6}{45.0} & \clsresult{73.0}{48.4} \\
QoQ-Med-7B$^\dagger$ & \clsresult{52.1}{30.6} & \clsresult{65.1}{62.2} & \clsresult{51.9}{30.5} & \clsresult{\textbf{57.9}}{25.8} & \clsresult{75.4}{27.2} & \clsresult{\textbf{72.0}}{41.8} & \clsresult{88.3}{46.9} \\
BreastGPT-8B & \clsresult{\textbf{73.6}}{\textbf{56.1}} & \clsresult{\textbf{78.6}}{\textbf{78.0}} & \clsresult{\textbf{70.6}}{\textbf{45.3}} & \clsresult{54.8}{\textbf{33.2}} & \clsresult{\textbf{89.3}}{\textbf{35.0}} & \clsresult{70.3}{\textbf{63.7}} & \clsresult{\textbf{89.5}}{\textbf{56.3}} \\
Lingshu-32B & \clsresult{59.9}{47.5} & \clsresult{67.9}{64.4} & \clsresult{62.8}{44.7} & \clsresult{23.0}{21.0} & \clsresult{45.2}{22.9} & \clsresult{65.4}{44.6} & \clsresult{78.5}{56.1} \\

\bottomrule
\end{tabularx}
\end{table*}
\begin{table*}[!htbp]
\centering
\scriptsize

\setlength{\tabcolsep}{3.6pt}
\renewcommand{\arraystretch}{1.30}

\setlength{\aboverulesep}{0pt}
\setlength{\belowrulesep}{0pt}

\caption{\textbf{Multiple-choice evaluation on BUS-BRA by task type.} Results are reported as \textbf{Accuracy} / \textcolor{iclrsubtext}{\textbf{Macro-F1}} (\%). $\dagger$ denotes thinking variants. Best results for each metric are shown in bold.}
\label{tab:app-choice-busbra}

\vspace{3pt}

\begin{tabularx}{0.60\linewidth}{
>{\raggedright\arraybackslash}p{0.18\linewidth}
*{3}{Y}
}

\toprule
\rowcolor{iclrheader}
&
\multicolumn{3}{c}{\cellcolor{iclrheader}\textbf{Assessment \& Diagnosis}}
\\

\noalign{%
    {\color{iclrheader}\hrule height \cmidrulewidth}%
    \vskip-\cmidrulewidth
}

\cmidrule(lr){2-4}

\rowcolor{iclrheader}
\multirow{-2}{*}{\textbf{Model}} & \textbf{AC} & \textbf{DX} & \textbf{PS}
\\
\midrule

\rowcolor{gray!10}
\multicolumn{4}{l}{\textbf{\textit{General-purpose VLMs}}}
\\
InternVL3-8B & \clsresult{25.7}{24.1} & \clsresult{41.8}{29.4} & \clsresult{47.1}{47.1} \\
Qwen2.5-VL-7B & \clsresult{37.0}{15.6} & \clsresult{32.8}{24.1} & \clsresult{47.4}{47.3} \\
Qwen3-VL-8B-I& \clsresult{35.3}{20.5} & \clsresult{41.1}{32.4} & \clsresult{60.2}{55.4} \\
Qwen3-VL-8B-T$^\dagger$ & \clsresult{31.2}{20.9} & \clsresult{32.2}{23.8} & \clsresult{56.9}{52.3} \\
Qwen2.5-VL-32B & \clsresult{34.1}{21.2} & \clsresult{37.8}{26.0} & \clsresult{58.6}{53.2} \\
Qwen3-VL-32B & \clsresult{36.2}{20.8} & \clsresult{43.2}{31.9} & \clsresult{55.8}{55.6} \\

\midrule
\rowcolor{gray!10}
\multicolumn{4}{l}{\textbf{\textit{Medical VLMs}}}
\\
MedGemma-4B & \clsresult{22.6}{22.4} & \clsresult{15.8}{15.1} & \clsresult{60.0}{51.8} \\
Lingshu-7B & \clsresult{38.6}{26.3} & \clsresult{58.1}{39.8} & \clsresult{73.3}{64.6} \\
Hulu-Med-7B & \clsresult{26.3}{25.4} & \clsresult{54.6}{45.0} & \clsresult{66.6}{61.8} \\
QoQ-Med-7B$^\dagger$ & \clsresult{36.9}{14.5} & \clsresult{39.9}{24.7} & \clsresult{35.4}{29.4} \\
BreastGPT-8B & \clsresult{\textbf{45.0}}{\textbf{36.5}} & \clsresult{\textbf{69.8}}{\textbf{54.4}} & \clsresult{\textbf{88.8}}{\textbf{87.1}} \\
Lingshu-32B & \clsresult{33.3}{28.0} & \clsresult{51.0}{28.7} & \clsresult{73.0}{68.1} \\

\bottomrule
\end{tabularx}
\end{table*}
\begin{table*}[!htbp]
\centering
\scriptsize

\setlength{\tabcolsep}{3.6pt}
\renewcommand{\arraystretch}{1.30}

\setlength{\aboverulesep}{0pt}
\setlength{\belowrulesep}{0pt}

\caption{\textbf{Multiple-choice evaluation on VinDr-Mammo by task type.} Results are reported as \textbf{Accuracy} / \textcolor{iclrsubtext}{\textbf{Macro-F1}} (\%). Best results for each metric are shown in bold.}
\label{tab:app-choice-vindr}

\vspace{3pt}

\begin{tabularx}{0.60\linewidth}{
>{\raggedright\arraybackslash}p{0.18\linewidth}
*{3}{Y}
}

\toprule

\rowcolor{iclrheader}
&
\textbf{Anatomy}
&
\textbf{Finding}
&
\textbf{Assessment \& Diagnosis}
\\

\noalign{%
    {\color{iclrheader}\hrule height \cmidrulewidth}%
    \vskip-\cmidrulewidth
}

\cmidrule(lr){2-2}
\cmidrule(lr){3-3}
\cmidrule(lr){4-4}

\rowcolor{iclrheader}
\multirow{-2}{*}{\textbf{Model}}
& \textbf{AA}
& \textbf{FT}
& \textbf{AC}
\\

\midrule

\rowcolor{gray!10}
\multicolumn{4}{l}{\textbf{\textit{General-purpose VLMs}}}
\\
InternVL3-8B & \clsresult{45.7}{23.8} & \clsresult{19.1}{8.7} & \clsresult{3.9}{4.2} \\
Qwen2.5-VL-7B & \clsresult{64.7}{26.6} & \clsresult{29.8}{10.2} & \clsresult{10.7}{9.6} \\
Qwen3-VL-8B-I& \clsresult{42.3}{26.2} & \clsresult{16.3}{9.0} & \clsresult{61.4}{22.0} \\
Qwen3-VL-8B-T$^\dagger$ & \clsresult{23.0}{14.9} & \clsresult{15.1}{7.3} & \clsresult{\textbf{63.4}}{19.9} \\
Qwen2.5-VL-32B & \clsresult{64.4}{28.0} & \clsresult{54.0}{12.7} & \clsresult{27.3}{15.9} \\
Qwen3-VL-32B & \clsresult{57.1}{28.0} & \clsresult{\textbf{54.6}}{13.3} & \clsresult{45.8}{20.1} \\

\midrule
\rowcolor{gray!10}
\multicolumn{4}{l}{\textbf{\textit{Medical VLMs}}}
\\
MedGemma-4B & \clsresult{18.3}{15.4} & \clsresult{23.2}{10.8} & \clsresult{11.8}{10.3} \\
Lingshu-7B & \clsresult{55.0}{28.3} & \clsresult{32.2}{12.3} & \clsresult{12.7}{11.3} \\
Hulu-Med-7B & \clsresult{36.2}{22.0} & \clsresult{17.0}{9.2} & \clsresult{21.2}{15.1} \\
QoQ-Med-7B$^\dagger$ & \clsresult{67.1}{24.7} & \clsresult{53.8}{12.0} & \clsresult{3.7}{2.5} \\
BreastGPT-8B & \clsresult{50.6}{\textbf{32.9}} & \clsresult{35.7}{\textbf{17.0}} & \clsresult{40.8}{\textbf{22.1}} \\
Lingshu-32B & \clsresult{\textbf{70.3}}{27.2} & \clsresult{42.7}{14.2} & \clsresult{45.8}{18.3} \\

\bottomrule
\end{tabularx}
\end{table*}
\begin{table*}[!htbp]
\centering
\scriptsize

\setlength{\tabcolsep}{3.6pt}
\renewcommand{\arraystretch}{1.30}

\setlength{\aboverulesep}{0pt}
\setlength{\belowrulesep}{0pt}

\caption{\textbf{Spatial evaluation on BUS-BRA and VinDr-Mammo.} Results report \textbf{mIoU} and $\mathbf{Acc@IoU}_{0.5}$ (\%). Best results for each metric are shown in bold.}
\label{tab:app-spatial-results}

\vspace{3pt}

\begin{tabularx}{0.80\linewidth}{
>{\raggedright\arraybackslash}p{0.18\linewidth}
*{6}{Y}
}

\toprule

\rowcolor{iclrheader}
&
\multicolumn{2}{c}{\cellcolor{iclrheader}\textbf{BUS-BRA Detection}}
&
\multicolumn{2}{c}{\cellcolor{iclrheader}\textbf{VinDr Detection}}
&
\multicolumn{2}{c}{\cellcolor{iclrheader}\textbf{VinDr Visual Grounding}}
\\

\noalign{%
    {\color{iclrheader}\hrule height \cmidrulewidth}%
    \vskip-\cmidrulewidth
}

\cmidrule(lr){2-3}
\cmidrule(lr){4-5}
\cmidrule(lr){6-7}

\rowcolor{iclrheader}
\multirow{-2}{*}{\textbf{Model}}
& \textbf{mIoU} & \textbf{Acc@.5}
& \textbf{mIoU} & \textbf{Acc@.5}
& \textbf{mIoU} & \textbf{Acc@.5}
\\

\midrule

\rowcolor{gray!10}
\multicolumn{7}{l}{\textbf{\textit{General-purpose VLMs}}}
\\

InternVL3-8B & 26.0 & 11.2 & 3.3 & 0.2 & 3.2 & 0.3 \\
Qwen2.5-VL-7B & 28.9 & 22.2 & 7.6 & 3.9 & 8.8 & 6.5 \\
Qwen3-VL-8B-I & 51.1 & 56.4 & 10.7 & 7.0 & 13.7 & 11.0 \\
Qwen3-VL-8B-T$^\dagger$ & 37.5 & 32.9 & 10.9 & 7.3 & 10.4 & 7.0 \\
Qwen2.5-VL-32B & 40.6 & 37.5 & 9.7 & 6.1 & 10.8 & 7.3 \\
Qwen3-VL-32B & 55.4 & 62.1 & \textbf{15.4} & \textbf{12.9} & \textbf{18.0} & \textbf{16.8} \\

\midrule

\rowcolor{gray!10}
\multicolumn{7}{l}{\textbf{\textit{Medical VLMs}}}
\\

MedGemma-4B & 15.7 & 3.7 & 1.5 & 0.0 & 1.6 & 0.0 \\
Lingshu-7B & 4.7 & 0.9 & 4.4 & 0.8 & 3.6 & 0.7 \\
Hulu-Med-7B & 7.0 & 3.0 & 0.0 & 0.0 & 0.0 & 0.0 \\
QoQ-Med-7B$^\dagger$ & 45.6 & 43.5 & 7.0 & 1.1 & 7.0 & 1.3 \\
BreastGPT-8B & \textbf{72.6} & \textbf{82.8} & 5.3 & 0.9 & 5.4 & 1.1 \\
Lingshu-32B & 12.5 & 4.4 & 4.7 & 1.5 & 5.6 & 2.0 \\

\bottomrule
\end{tabularx}
\end{table*}

The results allow us to examine how aggregate model performance is composed and where model
comparisons change with the evaluation target, metric, or imaging setting. We first analyze variation
within task families, then examine changes across model scales and thinking variants, and finally
compare categorical and spatial performance across imaging settings. Throughout, comparisons refer
to the evaluated checkpoints under the reported protocols.

\subsubsection{Task Granularity and Metric-Dependent Rankings}

Strong task-family performance can coexist with substantial variation among its constituent prediction targets. BreastGPT-8B achieves the highest aggregate finding-attribute scores, yet its BrEaST Macro-F1 ranges from 78.0 on margin and 63.7 on halo to 33.2 on posterior features and 35.0 on calcifications (Table~\ref{tab:app-choice-breast-attributes}). The aggregate FA score therefore combines markedly different attribute-level results. Retaining these prediction targets separately identifies which attributes contribute to a model's overall advantage and which remain comparatively weak under their respective evaluation settings.

The choice of metric also changes comparisons within an individual attribute. On posterior features, QoQ-Med-7B achieves higher Accuracy than BreastGPT-8B (57.9 versus 54.8), whereas BreastGPT-8B achieves higher Macro-F1 (33.2 versus 25.8). The same ordering reversal occurs on halo: QoQ-Med-7B leads in Accuracy (72.0 versus 70.3), but BreastGPT-8B leads in Macro-F1 (63.7 versus 41.8). These reversals extend the metric-dependent rankings observed in the main results to atomic prediction targets. Accuracy weights individual items equally, while Macro-F1 averages class-specific F1 scores; consequently, the two metrics capture different aspects of performance over the answer space. Reporting both preserves distinctions that either metric alone would omit.

\subsubsection{Changes across Model Scales and Thinking Variants}

Similar aggregate scores can conceal large changes in the underlying task profile. Qwen3-VL-8B-T exceeds Qwen3-VL-8B-I by only 0.60 Accuracy points overall, but the task-level changes are much larger (Table~\ref{tab:main_results}). FA improves by 27.09 points, while AA, FT, AC, DX, and PS decrease by 9.37, 1.22, 0.53, 4.85, and 7.50 points, respectively. Under the reported equal weighting of task families, the FA improvement contributes approximately 4.52 points to Overall Accuracy, while the other five tasks jointly contribute approximately $-3.91$ points. The small net change thus results from opposing task-level changes rather than uniformly similar performance.

The FA improvement is itself concentrated in particular attributes. On BrEaST, halo Accuracy increases from 11.4 to 63.4 and skin-thickening Accuracy from 2.7 to 76.6, whereas posterior-feature Accuracy decreases from 37.3 to 22.2 (Table~\ref{tab:app-choice-breast-attributes}). With seven equally weighted attributes, the halo and skin-thickening increases together contribute approximately 18.0 points to the roughly 27.1-point FA improvement, based on the rounded attribute scores. Decomposing the comparison at both levels therefore reveals a pattern that the Overall score obscures: a substantial gain within one task family is driven largely by a subset of its prediction targets and offset by declines elsewhere.

Changes across model scales also vary by task. Within Lingshu, the 32B checkpoint improves Overall Accuracy from 47.00 to 50.73 and FA Macro-F1 from 38.66 to 43.01 relative to the 7B checkpoint, while DX Macro-F1 decreases from 29.29 to 24.54. In contrast, Qwen3-VL-32B improves over Qwen3-VL-8B-I on aggregate multiple-choice performance and all three reported breast-imaging spatial settings (Tables~\ref{tab:main_results} and~\ref{tab:spatial_results}). These checkpoint comparisons show that aggregate improvements can have different task-level compositions. They support examining the resulting performance profile rather than using model scale or the thinking designation as a substitute for task-specific evaluation.

\subsubsection{Performance across Imaging Settings and Spatial Tasks}

Relative performance on categorical tasks does not consistently carry over to bounding-box prediction. On ISIC, Hulu-Med-7B achieves the highest diagnosis Accuracy and Macro-F1 among the evaluated checkpoints (56.3 and 34.4), but obtains 0.7 mIoU and 0.2 $\mathrm{Acc@IoU}_{0.5}$ on detection. Qwen3-VL-8B-I achieves lower diagnosis scores (53.2 and 22.8), yet substantially higher detection scores (80.3 and 91.4; Table~\ref{tab:app-isic-model-results}). These results describe complementary task profiles within the same dataset family. Because diagnosis and detection use different ISIC annotation subsets, the comparison concerns dataset-level performance rather than paired success and failure on individual images. The spatial scores measure end-to-end box prediction, including response validity under the evaluation protocol.

Spatial rankings also change across imaging settings. BreastGPT-8B leads Qwen3-VL-32B on BUS-BRA detection, with $\mathrm{Acc@IoU}_{0.5}$ of 82.80 versus 62.08, whereas Qwen3-VL-32B leads on VinDr-Mammo detection, with 12.86 versus 0.90 (Table~\ref{tab:spatial_results}). This ranking reversal shows that a model's advantage within one localization setting need not persist in another. Reporting each dataset--task pair separately preserves this distinction and prevents a pooled spatial score from obscuring where the observed advantage applies.

\end{document}